\pdfoutput=1
\documentclass[12pt, a4paper, oneside]{report}
\usepackage{float}												% customised float support
\usepackage{fancyhdr}											% improved header and footer support
\usepackage{amssymb}												% American Mathematical Society symbols
\usepackage{amsmath}												% American Mathematical Society environments
\usepackage{makeidx}												% automatic index generation
\usepackage{url}													% URL typesetting
\usepackage{texnames}											% BIBTeX, SliTeX, AMSTeX, PiCTeX and TeXsis logos
\usepackage{verbatim}											% multi-line comments
\usepackage{ifpdf}												% selective setup for PDF compilation
\usepackage[number=none]{glossary}							% glossary file environments

\usepackage[table,xcdraw]{xcolor}

\usepackage{caption}
\usepackage{subcaption}
\definecolor{carmine}{rgb}{0.59, 0.0, 0.09}
\definecolor{caputmortuum}{rgb}{0.35, 0.15, 0.13}
\definecolor{caputmortuum}{rgb}{0.35, 0.15, 0.13}
\definecolor{ao}{rgb}{0.0, 0.0, 1.0}
\definecolor{black}{rgb}{0.0, 0.0, 0.0}

\usepackage{xparse}
\usepackage{placeins}

\ifpdf

\usepackage{lineno}

\usepackage[
	bookmarks=true,												% (option) build bookmarks
	bookmarksnumbered=true,										% (option) add section numbers to bookmarks
	breaklinks
]{hyperref}

\usepackage{pdfpages}											% required for PDF watermarking
\usepackage{epstopdf}											% automatic conversion of EPS images
\usepackage[pdftex]{thumbpdf}									% thumbnail generation for PDF files
\usepackage{pdflscape}											% required by thumbpdf
\usepackage[all]{hypcap}										% correct PDF figure links to top of image

\hypersetup{
	pdfhighlight=/N,												% (option) no visual cue on clicking link
	pdffitwindow=true,											% (option)
	pdfstartview=Fit,												% (option) fit initial view to page
	plainpages=false,												% (option) prevent hyperref page number changes
	breaklinks=true,												% (option) allow link breaking across lines
	colorlinks=true,												% (option) color link text only (no borders)
	pageanchor=false,												% (option) turns off page referencing for title page
	linkcolor=blue,												% (option) internal link color
	citecolor=blue,												% (option) citation link color
	filecolor=blue,												% (option) file link color
	menucolor=blue,												% (option) Acrobat menu link color
	pagecolor=blue,												% (option) page link color
	urlcolor=blue,													% (option) URL link color
}

\hypersetup{
	pdftitle    = {A Replication Study: Machine Learning Models Are Capable of Predicting Sexual Orientation From Facial Images},
	pdfauthor   = {John Leuner <john.leuner@gmail.com>}
}

\else

\usepackage{graphicx}											% EPS graphics

\fi

\ifpdf
\else
\fi

\newlength{\originalbaselineskip}
\renewcommand{\gloskip}{}
\ifpdf
	
\else
	
\fi

\makeglossary
\makeindex

\newfloat{graph}{tbp}{lgf}[chapter]
\floatname{graph}{Graph}
\newfloat{algorithm}{tbp}{loa}[chapter]
\floatname{algorithm}{Algorithm}

\makeatletter
\long\def\@makecaption#1#2{%
	\vskip\abovecaptionskip
	\sbox\@tempboxa{{\small{\bf #1:} #2}}%
	\ifdim \wd\@tempboxa >\hsize
		{\small{\bf #1:} #2\par}
	\else
		\hbox to\hsize{\hfil\box\@tempboxa\hfil}%
	\fi
	\vskip\belowcaptionskip
}
\renewcommand\floatc@plain[2]{
	\setbox\@tempboxa\hbox{\small{\bf #1:} #2}%
	\ifdim\wd\@tempboxa>\hsize
		{\small{\bf #1:} #2\par}
	\else
		\hbox to\hsize{\hfil\box\@tempboxa\hfil}\fi}
\makeatother

\newlength{\abovesubfiglabelskip}
\begin{document}

%%%%%%%%%%%%%%%%%%%%%%%%%%%%%%%%%%%%%%%%%%%%%%%%%
%%%             Glossary Acronyms             %%%
%%%%%%%%%%%%%%%%%%%%%%%%%%%%%%%%%%%%%%%%%%%%%%%%%

\newacronym{AI}{Artificial Intelligence}{name=AI, description=Artificial Intelligence}
\newacronym{ANN}{Artificial Neural Network}{name=ANN, description=Artificial Neural Network}
\newacronym{CI}{Computational Intelligence}{name=CI, description=Computational Intelligence}

% and any other acronyms you want to use...

%%%%%%%%%%%%%%%%%%%%%%%%%%%%%%%%%%%%%%%%%%%%%%%%%
%%%           Index Sub-References            %%%
%%%%%%%%%%%%%%%%%%%%%%%%%%%%%%%%%%%%%%%%%%%%%%%%%

\index{AI|see{Artificial Intelligence}}
\index{ANN|see{Artificial Neural Network}}
\index{CI|see{Computational Intelligence}}

%%%%%%%%%%%%%%%%%%%%%%%%%%%%%%%%%%%%%%%%%%%%%%%%%
%%%%%%%%%%%%%%%%%%%%%%%%%%%%%%%%%%%%%%%%%%%%%%%%%

%%%%%%%%%%%%%%%%%%%%%%%%%%%%%%%%%%%%%%%%%%%%%%%%%
%%%%%%%%%%%%%%%%%%%%%%%%%%%%%%%%%%%%%%%%%%%%%%%%%
\setcounter{tocdepth}{1}

\newlength{\negativetitlepageoffset}
\setlength{\negativetitlepageoffset}{-5cm}

\begin{titlepage}
	\ \vspace{\negativetitlepageoffset}
	\vspace{\stretch{1}}
	\setlength{\baselineskip}{2.4\originalbaselineskip}
	\begin{center}
		\textsf{\huge A Replication Study: Machine Learning Models Are Capable of Predicting Sexual Orientation From Facial Images}
	\end{center}
	\begin{center}
		\textsf{by}
	\end{center}
	\begin{center}
		\textsf{\large John Leuner}
	\end{center}
	\vspace{\stretch{1}}
	\setlength{\baselineskip}{1.3\originalbaselineskip}
	\begin{center}
		\textsf{Submitted in partial fulfillment of the requirements for the degree\\
		Masters in Information Technology (Big Data Science)\\
		in the Faculty of Engineering, Built Environment and Information Technology\\
		University of Pretoria, Pretoria}
	\end{center}
	\vspace{0.15cm}
	\centerline{
		\textsf{November 2018}
	}
\end{titlepage}

%%%%%%%%%%%%%%%%%%%%%%%%%%%%%%%%%%%%%%%%%%%%%%%%%
%%%%%%%%%%%%%%%%%%%%%%%%%%%%%%%%%%%%%%%%%%%%%%%%%

\pagestyle{empty}
\newpage
\textsf{\small
	\vfill
	\noindent Publication data:\\*[2.5mm]
	\parbox{\textwidth}{
		\fontsize{9}{10pt}
		\selectfont
		John Leuner. A Replication Study: Machine Learning Models Are Capable of Predicting Sexual Orientation From Facial Images. Masters in Information Technology (Big Data Science) Dissertation, University of Pretoria, Department of Computer Science, Pretoria, South Africa, November 2018.
	}\\*[10.5mm]
	Electronic, hyperlinked versions of this dissertation are available online, as Adobe PDF files, at:\\*[2.5mm]
	\parbox{\textwidth}{
		\fontsize{9}{9.5pt}
		\selectfont
		\url{http://upetd.up.ac.za/UPeTD.htm}
	}
}

%%%%%%%%%%%%%%%%%%%%%%%%%%%%%%%%%%%%%%%%%%%%%%%%%
%%%%%%%%%%%%%%%%%%%%%%%%%%%%%%%%%%%%%%%%%%%%%%%%%

\newpage

\begin{center}
	{\large\bf A Replication Study: Machine Learning Models Are Capable of Predicting Sexual Orientation From Facial Images}
\end{center}
\begin{center}by\end{center}
\begin{center}
	{John Leuner}\\
	\ifpdf
		E-mail: \href{mailto:john.leuner@gmail.com}{john.leuner@gmail.com}
	\else
		E-mail: john.leuner@gmail.com
	\fi
\end{center}

\vspace{1cm}
\begin{center}{\large\bf Abstract}\end{center}
%\linenumbers
Recent research used machine learning methods to predict a person’s sexual orientation from their photograph (Wang and Kosinski, 2017). To verify this result, two of these models are replicated, one based on a deep neural network (DNN) and one on facial morphology (FM). Using a new dataset of 20,910 photographs from dating websites, the ability to predict sexual orientation is confirmed (DNN accuracy male 68\%, female 77\%, FM male 62\%, female 72\%). To investigate whether facial features such as brightness or predominant colours are predictive of sexual orientation, a new model trained on highly blurred facial images was created. This model was also able to predict sexual orientation (male 63\%, female 72\%). The tested models are invariant to intentional changes to a subject’s makeup, eyewear, facial hair and head pose (angle that the photograph is taken at). It is shown that the head pose is not correlated with sexual orientation. While demonstrating that dating profile images carry rich information about sexual orientation these results leave open the question of how much is determined by facial morphology and how much by differences in grooming, presentation and lifestyle. The advent of new technology that is able to detect sexual orientation in this way may have serious implications for the privacy and safety of gay men and women.

\vfill
\noindent
{\bf\parbox{26.8mm}{Supervisors}:} A. Bosman \\* % only provide titles of Prof. or Dr. (not Mr.)
{\bf\parbox{28.55mm}{~}} C. Stallmann \\*
%{\bf\parbox{26.8mm}{Supervisor}:} Prof.~S. U. P. Visor \\* % if you only have a single supervisor
{\bf\parbox{26.8mm}{Department}:} Department of Computer Science \\*
{\bf\parbox{26.8mm}{Degree}:} Master of Information Technology (Big Data Science)

%%%%%%%%%%%%%%%%%%%%%%%%%%%%%%%%%%%%%%%%%%%%%%%%%
%%%%%%%%%%%%%%%%%%%%%%%%%%%%%%%%%%%%%%%%%%%%%%%%%

\newpage
%
%\ \vspace{\stretch{1}}
%
%\begin{quotation}
%``An interesting quotation (preferably related to the theme of your research) if you would like to include one. If you choose not to include a quotation (or can't find anything relevant), remove this page.''
%\end{quotation}
%\begin{flushright}
%Quote attribution or source (1892)
%\end{flushright}
%
%\vspace{1cm}
%
%\begin{quotation}
%``Another quote, if you feel like it\ldots''
%\end{quotation}
%\begin{flushright}
%Another Quote attribution or source (1890)
%\end{flushright}
%
%\ \vspace{\stretch{1}}
%
%%%%%%%%%%%%%%%%%%%%%%%%%%%%%%%%%%%%%%%%%%%%%%%%%%
%%%%%%%%%%%%%%%%%%%%%%%%%%%%%%%%%%%%%%%%%%%%%%%%%%
%
%\newpage
%
%\begin{center}{\Large\bf Acknowledgements}\end{center}
%
%\vspace{0.3cm}
%
%\noindent If you wish to include any acknowledgements to anyone you feel was instrumental in the completion of the dissertation (or your continued survival through it's completion):
%\begin{itemize}
%	\item First person (or institution) you'd like to thank, and reasons;
%
%	\item Second person (or institution), and reasons;
%
%	\item Final person (or institution), and reasons.
%\end{itemize}

%%%%%%%%%%%%%%%%%%%%%%%%%%%%%%%%%%%%%%%%%%%%%%%%%
%%%%%%%%%%%%%%%%%%%%%%%%%%%%%%%%%%%%%%%%%%%%%%%%%

\cleardoublepage
\pagestyle{plain}
\pagenumbering{roman}
\setcounter{page}{1}
\ifpdf
\pdfbookmark[0]{Contents}{contents}
\fi
\tableofcontents

%%%%%%%%%%%%%%%%%%%%%%%%%%%%%%%%%%%%%%%%%%%%%%%%%
%%%%%%%%%%%%%%%%%%%%%%%%%%%%%%%%%%%%%%%%%%%%%%%%%

%TODO put back list of figures if necessary

\iffalse
\cleardoublepage
\ifpdf
\phantomsection
\fi
\addcontentsline{toc}{chapter}{List of Figures}
\listoffigures

%%%%%%%%%%%%%%%%%%%%%%%%%%%%%%%%%%%%%%%%%%%%%%%%%
%%%%%%%%%%%%%%%%%%%%%%%%%%%%%%%%%%%%%%%%%%%%%%%%%

\cleardoublepage
\ifpdf
\phantomsection
\fi
\addcontentsline{toc}{chapter}{List of Graphs}
\listof{graph}{List of Graphs}

%%%%%%%%%%%%%%%%%%%%%%%%%%%%%%%%%%%%%%%%%%%%%%%%%
%%%%%%%%%%%%%%%%%%%%%%%%%%%%%%%%%%%%%%%%%%%%%%%%%

\cleardoublepage
\ifpdf
\phantomsection
\fi
\addcontentsline{toc}{chapter}{List of Algorithms}
\listof{algorithm}{List of Algorithms}

%%%%%%%%%%%%%%%%%%%%%%%%%%%%%%%%%%%%%%%%%%%%%%%%%
%%%%%%%%%%%%%%%%%%%%%%%%%%%%%%%%%%%%%%%%%%%%%%%%%

\cleardoublepage
\ifpdf
\phantomsection
\fi
\addcontentsline{toc}{chapter}{List of Tables}
\listoftables

%%%%%%%%%%%%%%%%%%%%%%%%%%%%%%%%%%%%%%%%%%%%%%%%%
%%%%%%%%%%%%%%%%%%%%%%%%%%%%%%%%%%%%%%%%%%%%%%%%%
\fi

%%%%%%%%%%%%%%%%%%%%%%%%%%%%%%%%%%%%%%%%%%%%%%%%%
%%%%%%%%%%%%%%%%%%%%%%%%%%%%%%%%%%%%%%%%%%%%%%%%%

\ifpdf
\hypersetup{pageanchor=true}									% (option) turn page referencing back on for chapters
\fi

\pagestyle{fancy}
\fancypagestyle{headings}{
	\fancyhead[RO,LE]{\thepage}
	\fancyhead[LO]{\sf\nouppercase{\leftmark}}
	\fancyhead[RE]{\sf\nouppercase{\rightmark}}
	\fancyfoot{}
	\renewcommand{\headrulewidth}{0.4pt}
}

%%%%%%%%%%%%%%%%%%%%%%%%%%%%%%%%%%%%%%%%%%%%%%%%%
%%%%%%%%%%%%%%%%%%%%%%%%%%%%%%%%%%%%%%%%%%%%%%%%%
\chapter{Introduction}
\label{chap:introduction}
\pagestyle{headings}
\pagenumbering{arabic}
\setcounter{page}{1}
\graphicspath{{chapters/introduction/figures/}}
\graphicspath{{./chapters/introduction/figures/}}
The ability to predict sexual orientation from facial images using machine learning (ML) techniques has serious consequences for gay men and women. To verify previous results by Wang and Kosinski (W\&K)  \cite{wang_kosinski}, a replication study is undertaken with an independent dataset.
\section{Motivation}
This type of research falls within the discipline of Social Psychology, which studies how people act, think and feel in the context of society. Part of the field is concerned with how humans perceive other humans. Some of these studies investigate how well humans are able to perceive \textit{unambiguous} features of another person (such as their gender or age) and others focus on \textit{ambiguous} features that are not readily perceived in daily life, such as a person's sexual orientation or their political affiliation \cite{political_aff}.

While there have been previous studies demonstrating that humans have some skill at guessing someone's sexual orientation from various types of information \cite{ambady_hallahan_conner}, W\&K were the first to utilize the availability of new deep learning algorithms and online data sources to show that ML models are able to predict someone's sexual orientation from a photograph of their face \cite{wang_kosinski}.

\begin{figure*}[!b]
	\begin{subfigure}{.5\textwidth}
		\centering
		\includegraphics{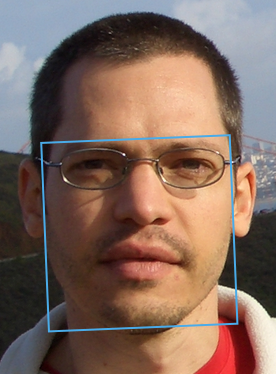}%
		\caption{The rectangle used to crop out the face.}
		\label{fig:intro_sfig1}
	\end{subfigure}%
	\begin{subfigure}{.5\textwidth}
		\centering
		\includegraphics{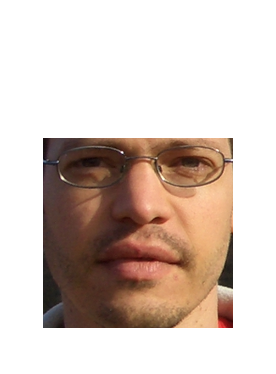}%
		\caption{Cropped facial image}
		\label{fig:intro_sfig2}
	\end{subfigure}
	\caption{The facial rectangle used to identify the face and the resulting cropped face used as an input by the deep learning ML models.}
	\label{fig:figure_fpp_crop}
\end{figure*}

\begin{figure*}[!b]
	\centering
	\centerline{
		\includegraphics[width=2.8in]{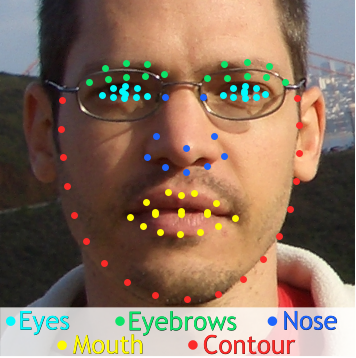}}%
	\caption{The facial morphology ``landmarks'' used as input by the facial morphology ML models.}%
	\label{fig:figure_fpp_example}%
\end{figure*}

Two of their models used ML techniques to predict sexual orientation from a photograph of a face. One used deep neural networks (DNN) \cite[Study 1a]{wang_kosinski} to extract features from the cropped facial image (see Figure \ref{fig:figure_fpp_crop}). The second study used only facial morphology \cite[Study 3]{wang_kosinski}. Facial morphology refers to the shape and position of the main facial features (such as eyes and nose) and the outline of the face (see Figure \ref{fig:figure_fpp_example} for an example of the information available in the facial morphology studies). The images were gathered from online dating profiles and from Facebook. The data was limited to white individuals from the United States.

For comparison, W\&K \cite{wang_kosinski} also ran an experiment in which they tested humans' ability to detect sexual orientation from the same photographs. They found that humans are able to predict sexual orientation with modest success, achieving an accuracy measured by the Area Under the Curve (AUC) of AUC=.61 for male images and AUC=.54 for female images. Both their ML models outperformed the humans. The deep learning classifier had an accuracy of AUC=.81 for males and AUC=.71 for females \cite[Study 1a]{wang_kosinski}. The facial morphology classifier scored AUC=.85 for males and AUC=.70 for females \cite[Study~4]{wang_kosinski}.
\newline
\indent W\&K note that their second model can predict sexual orientation from the individual components of the face, such as the contour of the face or the mouth. In particular, the facial contour predicts sexual orientation with an accuracy of AUC=.75 for men and AUC=.63 for women. Based on this outcome, W\&K make the claim that this validates a theory of sexual orientation called prenatal hormone theory (PHT) \cite{allen_gorski, jannini, udry}. PHT predicts that gay people will have ``gender-atypical'' facial morphology due to their exposure to differing hormone environments in the womb. They argue that this finding is unlikely to be due to different styles of presentation or grooming because it is quite difficult to alter the contour of one's face \cite{wang_kosinski}.

The theory that one can detect sexual orientation from facial differences due to a biological cause has been challenged by Ag{\"u}era y Arcas {\it et al} \cite{stereotypes}, who claim that there are several plausible explanations for these facial differences which do not rely on a biological origin. They note that there are many visual signals expressed through the face which are indicative of sexual orientation, such as the presence of eyewear, facial hair and the use of makeup. To get further insight into possible differences in presentation between gay and straight people, they built a questionnaire that asked the participants about their lifestyle and grooming habits. They found that only eight yes/no questions were sufficient to determine a person's sexual orientation with an accuracy of AUC=.70 for females \cite{stereotypes}.
\newpage
\indent Ag{\"u}era y Arcas {\it et al} proceed to question whether pre-natal hormonal exposure is responsible for facial differences in photographs. They argue that the angle that a photograph is taken at, especially when the person is taking a picture of themselves, has the effect of ``enlarging the chin, shortening the nose, shrinking the forehead, and attenuating the smile''. These differences could possibly result in either a deep learning classifier or a facial morphology classifier learning sexual orientation from differences due to the angle of the photograph and not any biological differences \cite{stereotypes}.

They go on to point out that there are obvious superficial differences between the gay and straight faces generated from W\&K's dataset (see \cite[Study 2]{wang_kosinski} for their composite images). Straight women in the composite image appear to wear more makeup than gay women and are less likely to wear glasses. Straight males are less likely to wear glasses and are more likely to have clearly visible facial hair \cite{stereotypes}.

To demonstrate how an alteration of appearance can make one appear ``more stereotypically gay'' or ``more stereotypically straight'', the researchers took photographs of themselves in which they altered their presentations \cite{stereotypes}. Figure \ref{fig:figure_altered_presentation} demonstrates these changes in appearance. The male photographs on the left are taken from a lower angle, have no eyewear and feature more facial hair. The female photograph on the left is shown with makeup and is taken from a higher angle. The photograph on the right has eyewear and is taken from a lower angle.
%\pagebreak
\thispagestyle{empty}
%\vspace*{\fill}

\noindent
\hspace*{\fill}
%\vspace{-\parskip}
%\vspace{-\abovedisplayskip}
\begin{minipage}[t]{0.8\textwidth}%\dimexpr0.5\linewidth}
	\centering
	\centerline{
		\includegraphics[width=2.3in]{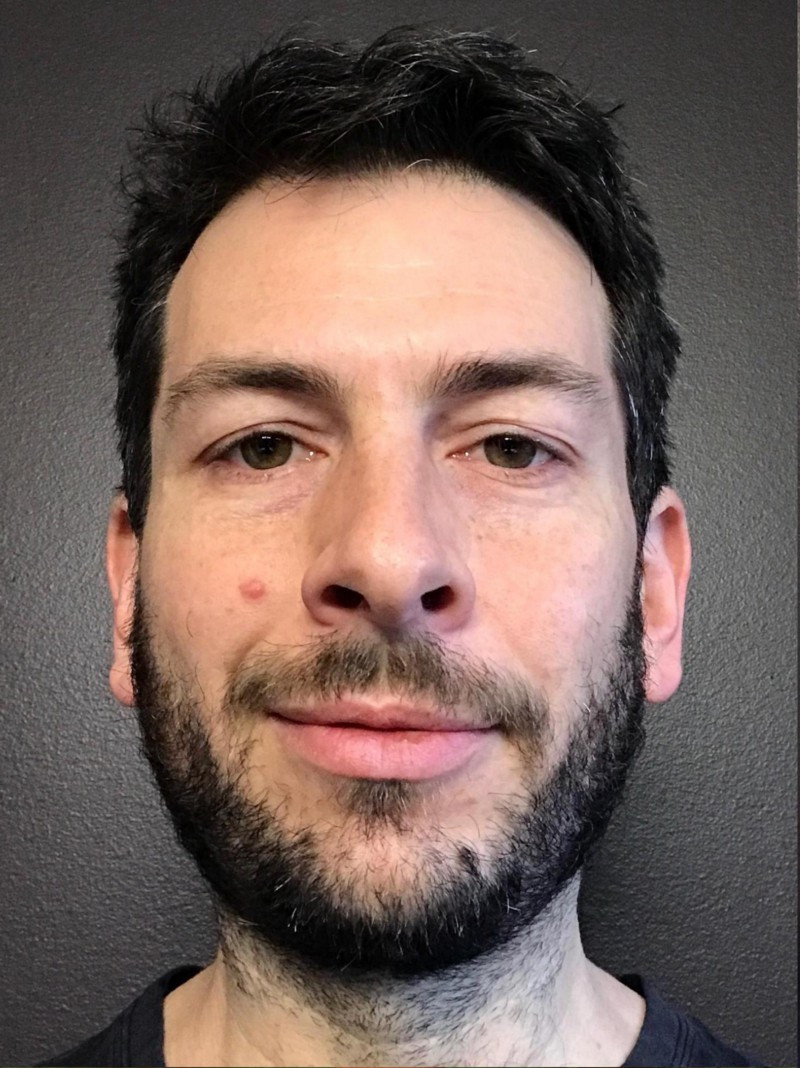}%
		\includegraphics[width=2.3in]{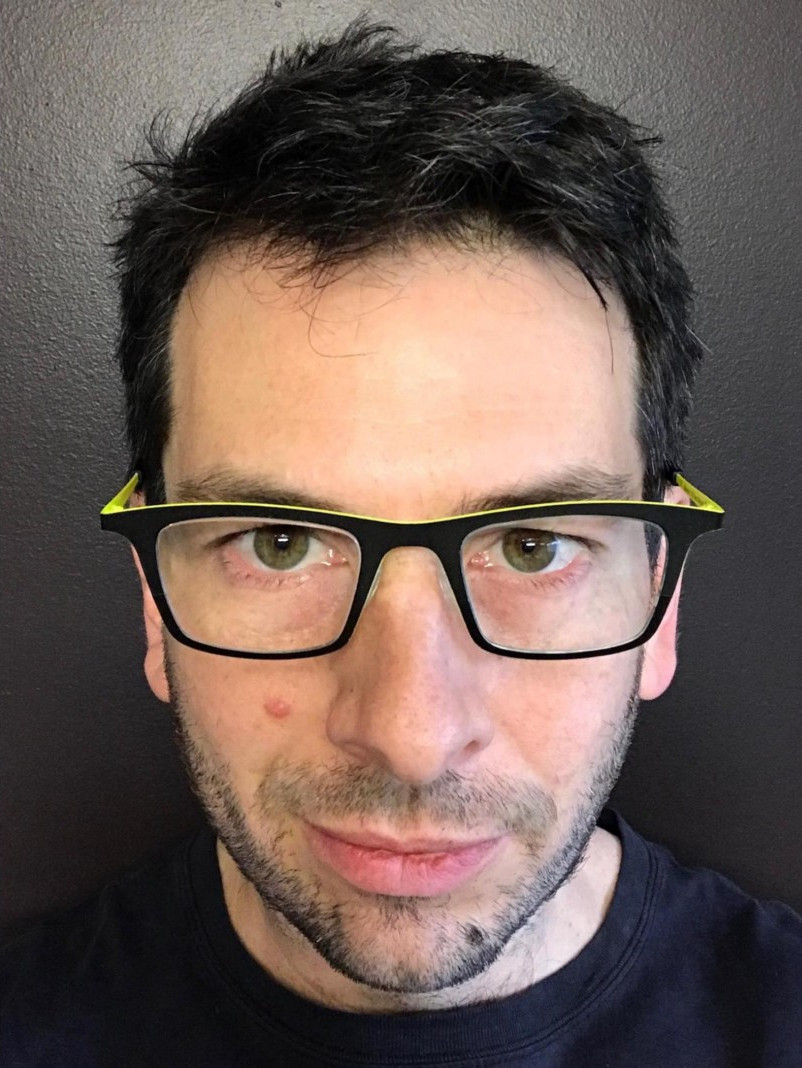}}%
	\centerline{
		\includegraphics[width=2.3in]{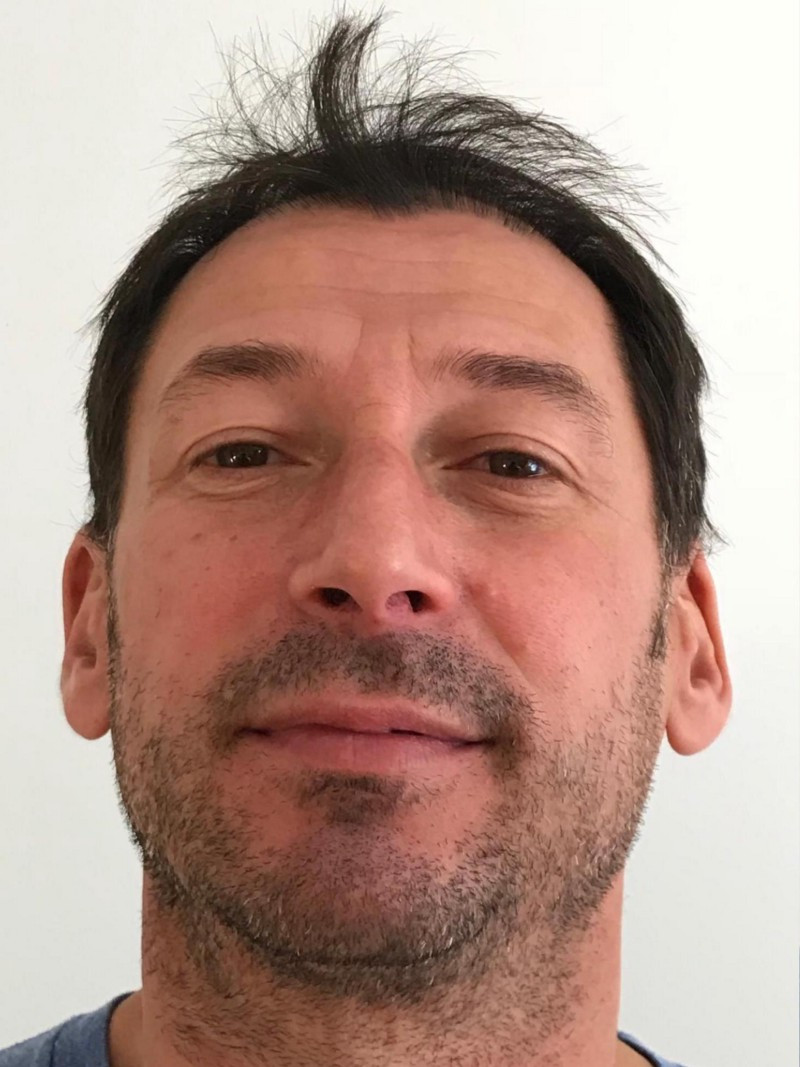}%
		\includegraphics[width=2.3in]{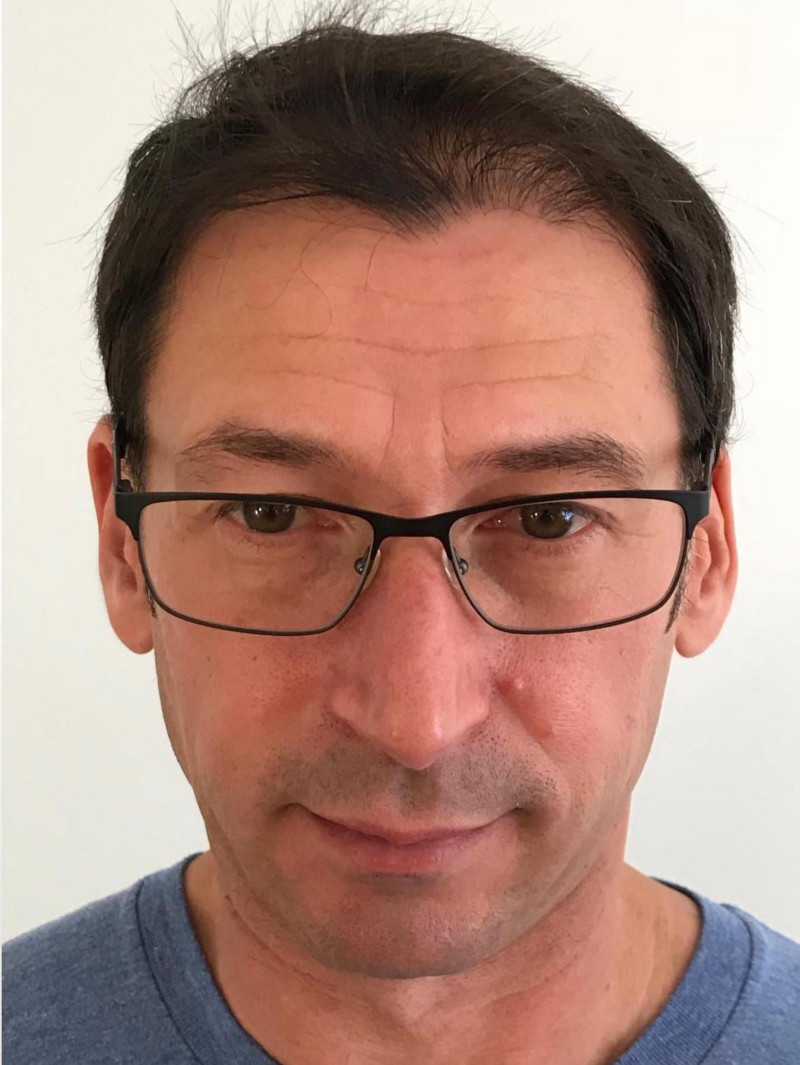}}%
	\centerline{
		\includegraphics[width=2.3in]{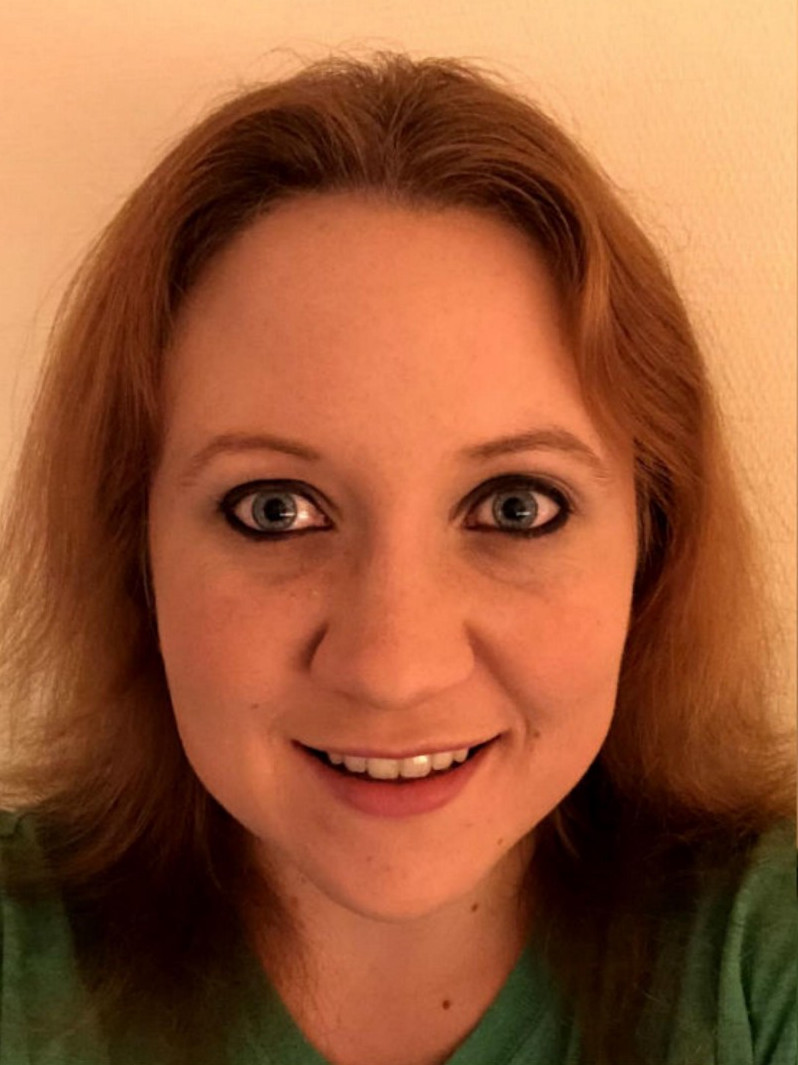}%
		\includegraphics[width=2.3in]{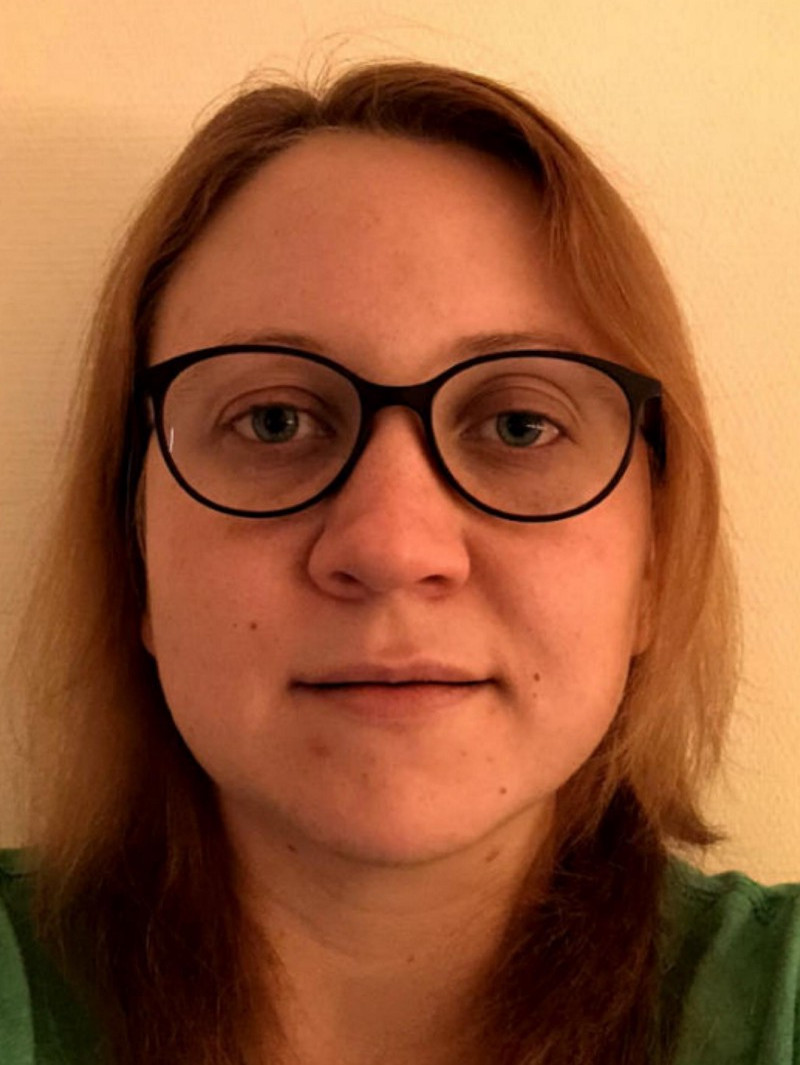}}%
	\captionof{figure}{Portraits of the same individual with altered presentation.}
	\label{fig:figure_altered_presentation}
\end{minipage}
\hspace*{\fill}
\vspace{\fill}
\pagebreak
\pagestyle{headings}
\section{Objectives}
The main objective of this study is to replicate the results of previous studies that used ML techniques to predict sexual orientation from facial images. Previous results show that ML techniques are able to do so better than humans can. These studies are repeated with a new dataset that is not controlled for race or country.

A new ML model based on highly blurred images is also introduced to investigate whether the information present in the predominant colours in the face and immediate background are predictive of sexual orientation.

The following factors are investigated to see whether the ML models are relying on them to make their predictions:
\begin{itemize}
	\item the \textit{head pose} (or angle at which the photograph is taken at),%
	\item the presence of facial hair or eyewear.%
\end{itemize}
\section{Contributions}
This work makes the following contributions in the area of social psychology:
\begin{itemize}
	\item The study replicates previous studies using ML techniques to predict sexual orientation from facial images. It is shown that both deep learning classifiers and facial morphology classifiers trained on photographs from dating profiles are able to predict sexual orientation, and that they do so better than humans can. The models use a new dataset not limited by race or country.
	
	\item The study introduces a new ML model that tests whether sexual orientation can be predicted from a highly blurred facial image. It is shown that the predominant colour information present in the face and background of a highly blurred facial image is predictive of sexual orientation.
	
	\item The study demonstrates that intentional alterations to one's appearance to fit gay and straight stereotypes do not alter the sexual orientation label generated by the ML models.
	
	\item The study shows that the \textit{head pose} is not correlated with sexual orientation.
	
	\item The study shows that the models are still able to predict sexual orientation even while controlling for the presence or absence of facial hair and eyewear.
\end{itemize}
\section{Thesis outline} 
The remainder of this thesis is organized as follows:
\begin{itemize}
	\item \textbf{Chapter \ref{chap:ethical}} discusses ethical issues related to this research
	
	\item \textbf{Chapter \ref{chap:literature}} reviews previous work
	
	\item \textbf{Chapter \ref{chap:dataset_ml_models}} describes the methods used to collect the dataset, and three ML models to predict sexual orientation:
	\begin{itemize}
		\item Dataset of facial images
		\item \textbf{Model 1}: Deep Neural Network (DNN) classifier
		\item \textbf{Model 2}: Facial morphology classifier
		\item \textbf{Model 3}: Highly blurred image classifier
	\end{itemize}
\item \textbf{Chapter \ref{chap:methods}} Describes the four experiments performed:
	\begin{itemize}
		\item \textbf{Study \hyperref[study:study1_ml]{1}}: Three ML models (Model~1, 2 and 3) are used to predict sexual orientation from facial images.
		\item \textbf{Study \hyperref[study:study2_altered_presentation]{2}}: Using the portraits in Figure \ref{fig:figure_altered_presentation}, ML models 1, 2 and 3 are evaluated to see whether alterations in presentation would result in changes to the predicted sexual orientation\footnote{Permission was granted by the authors to reproduce their photographs and to test the models on them.}.
		\item \textbf{Study \hyperref[study:study3_headpose]{3}}: Tests whether \textit{head pose} (or the angle that a photograph is taken at) is correlated with sexual orientation.
		\item  \textbf{Study \hyperref[study:study4_facial_hair_eyewear]{4}}: Tests whether the classifiers are still able to predict sexual orientation while controlling for two superficial features: facial hair or eyewear.
	\end{itemize}
\item \textbf{Chapter \ref{chap:results}} presents the results
\item \textbf{Chapter \ref{chap:discussion}} discusses the results and compares the findings to W\&K.
\end{itemize}
\chapter{Ethical Considerations}
\label{chap:ethical}
This chapter discusses some of the ethical issues that arise from using ML to predict people's sexual orientation. Section~\ref{ethics:rationale} discusses risks to privacy. Section~\ref{ethics:abuses} gives brief examples of historical abuses by scientists studying bodily features. Section~\ref{ethics:confidentiality} contains a note about the confidentiality of the data used in this study.
\section{Rationale for study}
\label{ethics:rationale}
After the publication of W\&K's work \cite{wang_kosinski} there was vociferous critique of their motives and their findings \cite{stereotypes, read_persons_character, glaad}. Todorov said that the implications of using AI\footnote{AI refers to Artificial Intelligence, which is equivalent to machine learning in this context.} to do ``face reading'' are ``morally abhorrent'' \cite{read_persons_character}. He argues that even if these algorithms are not capable of ``reading'' attributes like intelligence, political orientation and criminal inclination from photographs, merely talking about the possibility might encourage repressive governments to try employ these techniques \cite{read_persons_character}. We would add that the very idea that a government is capable of invading one's privacy in this way is in itself enough to create fear in targeted groups.
\subsection{Risks to privacy}
Kosinski argues that the world should be made aware of these techniques and the danger they pose. It has been demonstrated that besides the information betrayed by our photographs, digital trails such as who we link to on social media are also predictive of sexual orientation \cite{jernigan}. Kosinski says that ``there is an urgent need for making policymakers, the general public, and gay communities aware of the risks that they might be facing already'' \cite{wang_kosinski}.

We agree that the new risks society faces as a result of public digital profiles is too great to ignore. It's important to know whether our digital profiles do or do not reveal information like our sexual orientation or political affiliation. It's also important for people to know how to protect themselves. Todorov argues that the new ML techniques are only exposing what humans can tell anyway \cite{read_persons_character}, but even if the machines are only as good as humans, there is a fundamental difference when these processes can be automated by machines. Automated systems have the ability to operate at a completely different scale and with high fidelity over extended periods of time.
\subsubsection*{Automated surveillance}
The main new risk today is of automated surveillance by organisations most interested in uncovering people's intimate traits. Although easily within reach of mass surveillance states like the United States and China \cite{surveillance, surveillance_china}, automated tools for surveillance can easily be built with off-the-shelf technology by small organisations today.
\subsubsection*{Risks to gay people}
In many countries the threat of a loss of privacy regarding intimate traits is very serious. In South Africa, for example, practices such as ``corrective rape'' pose a grave threat~\cite{corrective_rape}. Worldwide there are at least eight countries in which homosexuality or homosexual acts are punishable by death \cite{death_penalty}.
\section{Historical abuses by scientists}
\label{ethics:abuses}
As described by W\&K \cite{wang_kosinski} there is a long history of the use of bodily features in a pseudo-scientific way to stigmatize, disenfranchise and marginalize people. In South Africa in particular there is a history of `race science' exemplified by the bodily measurements of 133 ``coloured'' males at the Zoology Department of Stellenbosch University in 1937 \cite{handri_walters}. These experiments ultimately led to the ``recognition of this population as a separate racial group that could be subjected to the laws of the apartheid state'' \cite{handri_walters}. In South Africa there is also a history of human rights abuses against homosexual people by health practitioners in academia during the apartheid era \cite{aversion}.
\subsubsection{Physiognomy}
Physiognomy is the reading of a person's character from their face \cite{read_persons_character}. This idea has a long history reaching back to the time of the ancient world. Today the scientific community rejects physiognomy as pseudo-science inspired by superstition and racism~\cite{jenkinson1997}. A recent example of physiognomy as pseudo-science is a Chinese study claiming to be able to detect criminality from identity photographs \cite{chinese_criminality}.
\section{Confidentiality and privacy in this study}
\label{ethics:confidentiality}
All of the data collected for this study is used to present results in aggregate only, and all data linked to individuals has been kept private. We have also not disclosed the sources of the data.
\iffalse
This study has been approved by the Faculty Committee for Research Ethics and Integrity (Faculty of Engineering, Built Environment \& Information Technology).
\fi
\chapter{Literature Review}
\label{chap:literature}
The study of sexual orientation and whether it can be perceived by others is not new. Section~\ref{lit:previous} briefly summarises previously published results. Section~\ref{lit:ml} describes the first use of ML to predict sexual orientation from a large dataset of photographs. It is not known how people or machines are able to predict sexual orientation from facial differences. Section~\ref{lit:biological} presents a claim for biological origins while Section~\ref{lit:rebuttal} contains a rebuttal of this idea.
\section{Previous work studying the prediction of sexual orientation}
\label{lit:previous}
Ambady {\it et al} \cite{ambady_hallahan_conner} in 1999 described how it is possible for humans to determine the sexual orientation of an individual through brief exposure to visual or audio clips with probability greater than chance. They summarise previous research which indicates that dynamic information (such as body movement, voice or video) provides more information than static features such as those present in a photograph. Their study finds that dynamic information is indeed superior to static information with the best accuracy of 70\% coming from silent video clips that are ten seconds long \cite{ambady_hallahan_conner}.

A later study by Rule and Ambady in 2007 showed that it is possible to perceive the sexual orientation of someone from a very brief exposure to a photograph of that person. By testing exposures differing in length of time they determined that an exposure as short as 50 ms is sufficient for a better than chance determination of sexual orientation. Results for longer exposures were not significantly different from the 50 ms exposure~\cite{brief_exposures}.

Olivola and Todorov \cite{olivola_2010b} reported in 2010 on the ability of website users to guess personal characteristics from a photograph. For most variables they found that people ignored the \textit{base rate} of the non-uniform distribution of that variable. For sexual orientation, they established via a survey that people assume that 90\% of people are heterosexual. However, when asked to predict sexual orientation from a photograph they ignored this \textit{base rate} and predicted a heterosexual sexual orientation 60\% of the time. Participants only achieved better than chance scores when the base rate approached equiprobability (50\%), for variables such as ``Is this person in a long term relationship?'' or ``Does this person have a college degree?''.

In 2013 Lyons also tested the ability of humans to predict sexual orientation from greyscale dating profile photographs and found that humans could do this with better than chance results \cite{lyons}.

In 2015 Cox {\it et al} \cite{coxdevine} found that human judges do not have a better than chance ability to predict sexual orientation from dating profile photographs.
\section{Use of machine learning to predict sexual orientation}
\label{lit:ml}
In 2017 a study by W\&K \cite{wang_kosinski} used static photographs of individuals collected from dating sites and Facebook to test whether humans could accurately judge sexual orientation from the photographs. They found that humans were able to achieve a 61\% success rate for classifying men \cite{wang_kosinski}. They then used a pre-trained deep learning neural network~\cite{deep_learning} on the same dataset and found that the resulting classifier was able to determine sexual orientation 81\% of the time for men and 71\% for women from a single image of the subject. When five images of the subject were used, the success rate increased to 91\% and 83\% respectively \cite[Study 1a]{wang_kosinski}.

The DNN they used is a pre-trained ML model that was originally built to identify someone from a photograph of their face. It was trained on a large number of photographs of celebrities. By modifying the model they were able to use the features it had already learned for a new task: predicting sexual orientation.

W\&K also used another pre-trained ML model \cite{faceplusplus} to derive the positions and shape of the main facial features (eyebrows, eyes, nose, mouth and facial contour). From these features they created a second model that also successfully predicted sexual orientation (male AUC=.85 and female AUC=.70 for five images per subject) \cite[Study 3]{wang_kosinski}.
\section{Claims of a biological origin for facial differences}
\label{lit:biological}
Placing emphasis on the ability of their second classifier to differentiate between gay and straight individuals based on the facial contour alone, W\&K claim that this result supports a theory called prenatal hormone theory (PHT) \cite{allen_gorski, jannini, udry}. PHT claims that same-gender sexual orientation in adults is a result of exposure to particular hormone environments before birth. W\&K claim that the theory predicts ``gender atypical'' faces for gay men and women (more feminine faces for gay men and more masculine faces for gay women) \cite{wang_kosinski}. 
\section{Rebuttal of Wang and Kosinski}
\label{lit:rebuttal}
Gelman {\it et al} point out that the type of classifier created by W\&K (and replicated in this study) is not useful in predicting sexual orientation for the general population because gay people are a minority and thus the classifier would have to be very accurate to be effective \cite{gelman}. They also criticize the use of binary categories (straight and gay) as an oversimplification of sexual orientation. To better communicate the uncertainty in the classifier results, they suggest that a third category should be reported when the classifier is unsure of the result \cite{gelman}.

Ag{\"u}era y Arcas {\it et al} \cite{stereotypes} responded to W\&K and provided alternative explanations for their findings. Instead of linking the ability to detect sexual orientation from facial images to a biological origin, they argue that ML models are learning from superficial features that have been hiding in plain sight \cite{stereotypes}. By inspecting the composite images of gay and straight males and females published by W\&K \cite[Study 1c]{wang_kosinski} they note the following apparent differences:
\begin{itemize}
	\item The composite straight female face has eyeshadow, the gay female face does not.
	\item Glasses (eyewear) are visible on both gay portraits but not the straight ones.
	\item Straight males appear to have more and darker facial hair.
	\item The gay male composite has a brighter face than the straight male and the straight female has a brighter face than the gay female composite.
\end{itemize}
To investigate whether there are any ``group preferences'' in grooming, presentation and lifestyle, they created a questionnaire with 77 questions asking respondents questions such as ``Do you wear eyeshadow?'',``Do you wear glasses?'', ``Do you have a beard?'' and other questions about gender and sexual orientation. Their survey polled 8000 Americans (the same demographic that W\&K used for their experiments).

They found that there is indeed a greater preference for wearing makeup and eyeshadow among the straight females polled relative to the gay females. Similarly, they found that gay males and females were more likely to report wearing glasses than straight men. In addition, they were also more likely to report that they like the way that they look in glasses.

Gay males reported being less likely to have ``serious facial hair'' than straight males (with the exception of gay males around 40 years old). Straight males also reported that they were more likely to ``work outdoors'' than gay males (which would explain less brightness in their faces).

All of these questionnaire responses match with the observations of the differences seen in W\&K's composite images. These apparent facial differences might then be the result of lifestyle, presentation and grooming preferences that differ between gay and straight groups.

The differences mentioned above are likely to be detected by a DNN such as the one used by W\&K for their first study. The second study, however, relies on facial morphology which should be insensitive to the kinds of differences described above. Ag{\"u}era y Arcas {\it et al} \cite{stereotypes} argue that the differences detected by the facial morphology classifier might simply be due to a relative difference in the pitch that a photograph is taken at.

Returning to the composite images, Ag{\"u}era y Arcas {\it et al} \cite{stereotypes} point out that the nostrils (which appear as dark ovals) are more prominent for photographs of straight males and gay females. They postulate that this might be due to a preference for taking pictures from a higher angle in the other two groups. Taking pictures from a higher angle also changes the shape of features such as the eyebrows and eyes, and this could easily be detected by a facial morphology classifier \cite{stereotypes}. 
\section{Summary}
This chapter reviewed prior work studying whether it is possible for humans to predict sexual orientation from photographs or other types of recording (such as videos). W\&K claim that sexual orientation is detectable from faces due to biological differences consistent with PHT. Ag{\"u}era y Arcas {\it et al} rebut this claim and argue that the facial differences are due to lifestyle choices and differences in presentation and grooming.
\chapter{Dataset and Machine Learning Models}
\label{chap:dataset_ml_models}
This chapter describes the dataset and machine learning models used in this study. Section~\ref{section_dataset} explains how the dating profile images were collected, cleaned and labelled. The remaining sections describe the three ML models. Section~\ref{model:vgg} describes Model~1, based on a deep neural network. Section~\ref{model:fpp} describes Model~2, based on facial morphology. Section~\ref{model:blurred} describes Model~3, trained on highly blurred images.
\section{Dataset of facial images}
\label{section_dataset}
\subsection{Dataset acquisition}
To replicate the experiments by W\&K \cite{wang_kosinski}, an independent dataset of facial images was collected from online dating sites using similar criteria.

Dating sites were used as a source for facial images because they are currently\footnote{in 2018} the most practical source of images for this kind of study. By stating their own gender and expressing an interest in another gender, dating profiles provide a convenient set of photographs labelled with a proxy for each person's sexual orientation.

Dating sites provide the opportunity to gather larger sample sizes than have previously been available to conventional studies which have to gather portraits individually, or manually extract them from a published source such as a newspaper.

For each dating profile retrieved, the following information was captured:
\begin{itemize}
	\item One or more photographs of that person
	\item The gender of the person
	\item The gender that they are seeking
	\item The age of the person
\end{itemize}
\subsubsection*{Terminology}
In this chapter and in the rest of this work we use the word `gay' to refer to someone who has expressed an interest in someone with the same gender (same sex attracted). We use the word `straight' to refer to someone who has expressed an interest in someone of a different gender (opposite sex attracted)\footnote{We use terminology suggested by the American Psychological Association \cite{avoiding_heterosexual_bias}}. The sources that we used to gather the data generally only allow people to identify themselves as belonging to one binary gender (male or female). We excluded users that expressed an interest in more than one gender.

Each photograph (facial image) is labelled with the gender of the subject and the gender of the person that they are seeking.

Every photograph then falls into one of four basic categories:
\begin{itemize}
	\item Gay Female
	\item Straight Female
	\item Gay Male
	\item Straight Male
\end{itemize}
The first dating site used to gather images, \textbf{Site A}, did not have a sufficient number of gay males and gay females. To increase the number of samples, images were collected from two additional sites. \textbf{Site B} caters to gay males only. \textbf{Site C} caters to gay females only. To protect the privacy of the subjects, the origin of the data sources are not disclosed.

In total about 500,000 photographs were retrieved. Table \ref{data_source_table1} lists the number of photographs downloaded in each category for each data source. 
\begin{table}[hb]
\centering
	\begin{tabular}{llll}
		\hline
		\rowcolor[HTML]{000000}
		\multicolumn{1}{|l|}{\cellcolor[HTML]{000000}{\color[HTML]{FFFFFF} }} & \multicolumn{1}{l|}{\cellcolor[HTML]{000000}{\color[HTML]{FFFFFF} \textbf{Site A}}} & \multicolumn{1}{l|}{\cellcolor[HTML]{000000}{\color[HTML]{FFFFFF} \textbf{Site B}}} & \multicolumn{1}{l|}{\cellcolor[HTML]{000000}{\color[HTML]{FFFFFF} \textbf{Site C}}} \\ \hline%
		Female - gay                                                     & 5951                                                                                &                                                                                     & 75768                                                                               \\ \hline%
		Female - straight                                                 & 292588                                                                              &                                                                                     &                                                                                     \\ \hline%
		Male - gay                                                       & 11030                                                                               & 104198                                                                              &                                                                                     \\ \hline%
		Male - straight                                                   & 72818                                                                               &                                                                                     &                                                                                     \\ \hline%
	\end{tabular}
	\caption{Number of photographs retrieved in each category from each data source}%
	\label{data_source_table1}%
\end{table}%
\subsection{Dataset cleaning and facial extraction}
For each photograph and associated user profile, a number of filtering and cleaning steps were applied. Only photographs for users aged 18 to 35 were included \footnote{This is the same age range used by W\&K\cite{wang_kosinski}}. Users that had missing or invalid data, such as missing age or gender labels, were removed.

Faces were automatically extracted from each photograph by using an external face recognition model called \textit{Face++}\footnote{Face++ can be accessed at \url{https://www.faceplusplus.com}}. \textit{Face++} generates a rectangle identifying the face (see Figure \ref{fig:figure_fpp_crop}) and facial metrics that describe facial ``landmarks" (see Figure \ref{fig:figure_fpp_example}). It also returns a \textit{head pose} for the face, comprised of the pitch, roll and yaw angles.

To avoid ambiguity, all photographs that did not contain exactly one face were discarded. Based on the facial metrics, photographs were excluded that:
\begin{itemize}
	\item Have a pitch of greater than 10 degrees or a yaw of greater than 15 degrees
	\item Have a distance between the eyes of less than 40 pixels
	\item Do not contain all of the facial landmarks normally detected by \textit{Face++}
\end{itemize}
These steps exclude images where the subject is not facing the camera, is not close to the camera or whose face is partially occluded.
\subsection{Manual filtering and labelling}
After the automatic filtering processes each image was inspected manually and rejected if it did not meet the following criteria:
\begin{itemize}
	\item Each image must be of a single adult person (often users upload pictures of animals or children).
	\item The image should be clear and not blurred or overexposed.
	\item The person in the image should match the category assigned to them (for example, verify that the picture is of a male).
	\item The photograph must not be digitally transformed or stylized in an obvious way, such as by the addition of digitally generated sunglasses or noses.
	\item The photograph must not be of a celebrity.
	\item The subject's face must not be obscured or occluded in the photograph\footnote{In some cases a mobile phone is visible in the lower half of the face when subjects are photographing themselves in a mirror}.
\end{itemize}
In addition, photographs were rejected if it was obvious that the user had mislabelled their own gender. This happens quite frequently, and we assume that it is because some users may not be computer literate (or literate in English). It is possible that they use the default gender setting when creating their account (although we observed this phenomenon for both males and females). This is different from people who purposely present themselves as belonging to a gender different to their sex assigned at birth.
\subsubsection*{Facial hair}
Each image was also labelled with an indicator marking the presence (or absence) of significant facial hair. For the purposes of this study, ``facial hair'' or ``significant facial hair'' is defined as either:
\begin{itemize}
	\item Clearly visible hair (several millimetres in length) in the moustache, beard or cheek areas
	\item Short stubble in the moustache, beard or cheek areas
\end{itemize}
The purpose of this definition is to test whether a grooming preference for facial hair is what is being recognized by the ML models and used to predict sexual orientation. None of the female subjects exhibited facial hair.
\subsubsection*{Eyewear}
Similarly, it was recorded whether the subject in each image was or was not wearing eyewear. In this study spectacles, sunglasses and mock or novelty spectacles are considered to qualify as eyewear. Images with digitally generated spectacles were discarded.
\subsubsection*{Filtering and labelling methodology}
To filter and label the photographs, software was developed to review the photographs. Photographs were randomly selected from each category for the reviewing process in batches of a hundred. Reviewers could not see the stated sexual orientation for each photograph. This process was repeated for the different categories until there were roughly 5000 reviewed images available for each category.

Each cropped face was presented to the user along with the stated gender of the subject. Users had the option to view the full photograph. For each photograph, an option to approve or reject the photograph was presented (with neither option selected by default). If approved, options to label the presence or absence of facial hair and eyewear were presented (none selected by default). A photograph was only approved if the facial hair and eyewear sections were completed.
\subsection{Final photograph count after automatic and manual filtering}
In Table \ref{data_source_table2}, the final number of photographs are shown in each category after the automatic filtering process and manual approval or rejection by a human.
\begin{table}[!ht]%
\centering
	\begin{tabular}{ll}
		\hline
		\rowcolor[HTML]{000000}%
		\multicolumn{1}{|l|}{\cellcolor[HTML]{000000}{\color[HTML]{FFFFFF} }} & \multicolumn{1}{l|}{\cellcolor[HTML]{000000}{\color[HTML]{FFFFFF} \textbf{Total}}} \\ \hline
		Female - gay                                                     & 5132                                                                               \\ \hline
		Female - straight                                                 & 5406                                                                               \\ \hline
		Male - gay                                                       & 5706                                                                               \\ \hline
		Male - straight                                                   & 4666                                                                               \\ \hline
	\end{tabular}%
	\caption{Total number of photographs reviewed and approved by a human in each category}%
	\label{data_source_table2}%
\end{table}
\section{Model 1: Deep neural network classifier}
\label{model:vgg}
We recreate the DNN model used by W\&K to predict sexual orientation from facial images \cite[Study 1a]{wang_kosinski}.
\subsection{VGGFace}
This model uses VGGFace \cite{vgg16}, a pre-trained deep learning neural network, to extract features from facial images, and then trains a logistic regression model on these features to predict the sexual orientation of the subject of the image.

VGGFace is a convolutional neural network that was developed to recognise individuals from pictures of their faces. It was trained on one million photographs of 2,622 different celebrities. Although the neural network was originally developed to identify a specific person from a facial image, by removing the last layer of the network we are able to obtain the facial features that the model uses for its final classification layer.

These features produced by a DNN are generally not interpretable by humans, but can be thought of as a numerical representation of a face.
\subsection{Model pipeline}
The input to the model is a cropped facial image extracted by \textit{Face++} \cite{faceplusplus}. See Figure~\ref{fig:figure_fpp_crop} for an example of a cropped face. The image is then scaled down to a 224 by 224 pixel image.

Next we used VGG16 \cite{vgg16}, a variant of VGGFace, to generate 4,096 features from each facial image.

We perform dimensionality reduction on the features using principal components analysis (PCA) \cite{pca}. Each image is represented by 500 principal components.

Each image is associated with a sexual orientation label (gay or straight) derived from an individual's reported gender and dating interest (see Section~\ref{section_dataset}).

To predict the sexual orientation of the individual in each image, we trained a logistic regression model using the principal components as independent variables and the sexual orientation labels as dependent variables. Males and females were modelled separately. This method is the same as that used by W\&K except that they used singular value decomposition (SVD) instead of PCA for dimensionality reduction \cite{wang_kosinski}.
\section{Model 2: Facial morphology classifier}
\label{model:fpp}
We recreate the facial morphology model used by W\&K  to predict sexual orientation from facial images \cite[Study 3]{wang_kosinski}.
\subsection{Face++}
This model uses \textit{Face++}\footnote{Face++ can be accessed at \url{https://www.faceplusplus.com}} \cite{faceplusplus}, an external model accessible as a service, to extract facial ``landmarks" for each face. It then uses distances derived from these landmarks to train a logistic regression model to predict the sexual orientation of the subject of the image.

The landmarks are facial metrics that describe the position of facial features on the face. \textit{Face++} returns a fixed number of landmark points for a face. The landmarks are grouped into several components:
\begin{itemize}
	\item[--] Contour (19 points)
	\item[--] Mouth (18 points)
	\item[--] Eyebrows (16 points)
	\item[--] Eyes (20 points)
	\item[--] Nose (10 points)
\end{itemize}
Figure \ref{fig:figure_fpp_example} demonstrates the landmarks for each facial component. Each component consists of ten or more points, and there are a total of 83 points for the whole face. Logistic regression classifiers were created for each individual facial component, and a combined classifier from all the components for the whole face.
\subsection{Model pipeline}
Each facial image is passed to \textit{Face++} to generate landmarks for the face. To generate features suitable for ML the same method as used by W\&K \cite{wang_kosinski} is followed. The distance from each point to every other point within each component is calculated. For example, since the nose has 10 points, $ 10 \times 9 = 90 $ euclidean distances are calculated for this component. All the distances are scaled by dividing them by the distance between the centre of the eyes.

We used PCA on the features. For the combined classifier which incorporates all the facial components, we reduced the number of components to 500.

Each image is associated with a sexual orientation label (gay or straight) derived from an individual's reported gender and dating interest (see Section~\ref{section_dataset}).

To predict the sexual orientation of the individual in each image, we trained a logistic regression model using the principal components as independent variables and the sexual orientation labels as dependent variables. Males and females were modelled separately. This method is the same as that used by W\&K except that they used singular value decomposition (SVD) instead of PCA for dimensionality reduction \cite{wang_kosinski}.
\section{Model 3: Highly blurred image classifier}
\label{model:blurred}
We create a new ML model to predict sexual orientation from facial images using a highly blurred image of the face.

The purpose of this model is to test whether the colour information alone is predictive of sexual orientation. The cropped and blurred image that is presented to the classifier contains no information about the shape or size of the facial features. This contrasts with the facial morphology classifier which relies solely on the shape of the facial features.
\newcommand{\plh}{%
	{\ooalign{$\phantom{0}$\cr\hidewidth$\scriptstyle\times$\cr}}%
}
\newcommand{\PLH}{{\mkern-2mu\times\mkern-2mu}}%
Two types of blurred image are created, the first is a 5$\plh$5 pixel image containing 25~colours, and the second is a 1 pixel image containing a single colour value.
\subsection{Blurring}
\begin{figure}[!h]
	\vspace{0.2in}
	\begin{subfigure}{.33\textwidth}
		\centering
		\includegraphics{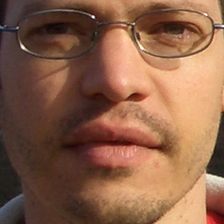}%
		\caption{Cropped facial image}%
		\label{fig:sfig1}%
	\end{subfigure}%
	\begin{subfigure}{.33\textwidth}
		\centering
		\includegraphics{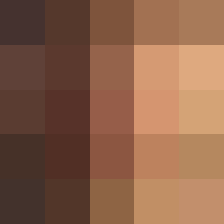}%
		\caption{$5\plh5$ pixel blurred image}%
		\label{fig:sfig2}%
	\end{subfigure}%
	\begin{subfigure}{.33\textwidth}
		\centering
		\includegraphics{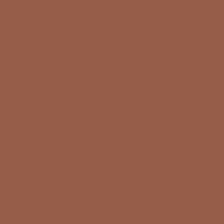}%
		\caption{1 pixel blurred image}%
		\label{fig:sfig3}%
	\end{subfigure}%
	\caption{The cropped facial image and derived blurred photos (blurred images are shown enlarged).}%
	\label{fig:figure_study_3_blurred_image}%
\end{figure}

\FloatBarrier
Cropped facial images (Figure \ref{fig:figure_fpp_crop}) are passed through a blurring algorithm\footnote{Images were blurred using ImageMagick's \textit{scale} operator \cite{imagemagick_scale_operator}.} to generate a highly blurred image. Figure \ref{fig:figure_study_3_blurred_image} shows an example of the original cropped facial image and the resulting blurred 5$\plh$5 pixel and 1~pixel images. The blurred images are shown enlarged because the actual 5$\plh$5 pixel and 1~pixel images are very small.

Since each pixel has three RGB colour channels, the resulting images have ($25~\times~3~=~75$) and ($1 \times 3 = 3$) features per image respectively.
\FloatBarrier
\subsection{Model pipeline}
Each cropped facial image is blurred to produce 75 or 3 features, for 5$\plh$5 pixel and 1~pixel images, respectively.

Each image is associated with a sexual orientation label (gay or straight) derived from an individual's reported gender and dating interest (see Section~\ref{section_dataset}).

To predict the sexual orientation of the individual in each image, we trained a logistic regression model using the features as independent variables and the sexual orientation labels as dependent variables. Males and females were modelled separately.
\section{Summary}
This chapter described the collection, cleaning and labelling process for the dataset of dating profile images. It also described the model pipelines for the three different ML models used in this study. The next chapter describes the methodology for the experiments performed using this dataset and these ML models.
\FloatBarrier
\chapter{Methodology}
\label{chap:methods}
Using the dataset and ML models from Chapter \ref{chap:dataset_ml_models}, four experiments are performed to investigate whether it is possible to predict sexual orientation from a facial image, and which properties of the facial images might make that possible.

Study \hyperref[study:study1_ml]{1} in Section~\ref{study:study1_ml} uses the three ML models to predict sexual orientation from facial images. Study \hyperref[study:study2_altered_presentation]{2} in Section~\ref{study:study2_altered_presentation} evaluates whether alterations to presentation result in changes to the predicted sexual orientation. Study \hyperref[study:study3_headpose]{3} in Section~\ref{study:study3_headpose} tests whether the \textit{head pose} (or the angle that a photograph is taken at) is correlated with sexual orientation. Study \hyperref[study:study4_facial_hair_eyewear]{4} in Section~\ref{study:study4_facial_hair_eyewear} controls for facial hair and eyewear as potential confounders while predicting sexual orientation.
\section{Study 1: Prediction of sexual orientation from facial images using machine learning models}
\label{study:study1_ml}
To replicate W\&K's ML studies \cite{wang_kosinski} we repeated their experiments using a deep learning classifier (Model~\hyperref[model:vgg]{1}) and a facial morphology classifier (Model~\hyperref[model:fpp]{2}) on an independent dataset (see Section~\ref{section_dataset}). We also created a new ML model that tests whether sexual orientation can be predicted from a highly blurred facial image (Model~\hyperref[model:blurred]{3}).

For all three models we use the same methods for training and scoring, discussed below.
\label{section:statistical_methods}
\subsubsection*{Model training}
Each base dataset contains samples which are labelled as being in either a positive or negative class. For example, in this study, the positive class might be ``Gay Male'' and the negative class ``Straight Male''. It doesn't matter which label is used as the positive class. To balance out the number of samples in each class, samples are randomly selected without replacement from each class in an alternate fashion until there are no samples left in one of the classes. The final dataset used for training thus has 50\% of its samples coming from the positive class and 50\% from the negative class.

To avoid overfitting, a stratified 20-fold cross validation technique is used. The data is randomly split into 20 parts, whilst preserving the 50/50 split between classes in each part. For each part a logistic regression model is trained on the remaining 19 parts and then used to generate a prediction score for every sample in that part.

The logistic regression model is configured to use L1 regularization \cite{lasso} with a default regression strength parameter C=1.0. A grid search was used to check that the default parameter performed best.
\subsubsection*{Model scoring}
\label{section:model_scoring}
The prediction score for each sample, which lies between 0 and 1, is used to predict whether the sample falls in the positive or the negative class. A value below the threshold of 0.5 places the sample in the positive class, all other values fall in the negative class. 

Using the predicted labels and the actual labels for each sample, a false positive rate and a true positive rate is calculated. These two values then allow us to calculate the area under the curve (AUC) of the receiver operating characteristic (ROC) \cite{roc_auc}.

The ROC AUC score represents the probability that when given one randomly chosen positive instance and one randomly chosen negative instance, the classifier will correctly identify the positive instance.

Note that the ROC AUC score used here is invariant to the threshold used for the classifer \cite{roc_auc}.
\subsubsection*{Accuracy for increasing numbers of images per subject}
Some of the subjects in the dataset have more than one facial image associated with them. To assess how the accuracy of the model changes with more than one image available per subject, we constructed accuracy scores for 1 to 5 images. To create these scores, we used the following procedure:

For $n$ from 1 to 5:
\begin{itemize}
	\item[--] Select all subjects with at least $n$ photos.
	\item[--] For each subject randomly select $n$ photos.
	\item[--] For each subject average the prediction scores for the photos.
	\item[--] Score the model by comparing the averaged predicted class per subject against their actual class label.
\end{itemize}
Prediction scores were logit transformed before averaging, and the reverse transformation was applied after averaging \cite{logit}. 

Finally, we have 5 scores representing the accuracy of the models when presented with $n$ images per subject. The scores for higher numbers of facial images vary due to the random sampling of photos in the scoring procedure, so we repeat it ten times and average the results.
\section{Study 2: Evaluate machine learning models with altered presentation}
\label{study:study2_altered_presentation}
To test whether an individual's change in presentation changes the predicted sexual orientation labels, we evaluated each pair of portraits in Figure \ref{fig:figure_altered_presentation} using the ML models tested in this study\footnote{Permission was granted by the authors to reproduce their photographs and to test the models on them.}.
\subsubsection*{Differences in presentation}
Each pair of photographs has a significant difference in pitch of more than 11 degrees. All the images on the left are taken without eyewear, while those on the right do have eyewear. The male portraits are shown with more facial hair on the left than on the right. The male portraits are taken from a low angle on the left and a high angle on the right. The female portrait on the left is shown with makeup and is taken from a high angle. The female portrait on the right has no makeup and is taken from a level angle.

These changes in presentation are in line with Ag{\"u}era y Arcas {\it et al}'s survey of 8000 Americans showing that straight males up to the age of 35 are more likely to report having serious facial hair than gay males. A preference for how they look in glasses and a higher prevalence of wearing glasses is reported by both gay males and gay females. Gay women report wearing makeup less often than straight women \cite{stereotypes}.

To test whether the models predict the same sexual orientation label for each pair of photographs we evaluated all six photographs using the three models from Chapter \ref{chap:dataset_ml_models}:
\begin{itemize}
	\item \textbf{Model 1}: Deep learning classifier 
	\item \textbf{Model 2}: Facial morphology classifier 
	\item \textbf{Model 3}: Blurred image classifier 
\end{itemize}
This produces 18 sexual orientation predictions, or 9 pairs of predictions.

We test the outcome to see if the predicted labels are pairwise consistent (does the model predict the same label for two images of the same individual?), and whether the labels are consistent between models (do the models agree on the predicted sexual orientation of an individual?).
\subsubsection*{Probability of pairwise consistency}
The probability of the models being pairwise consistent by chance is calculated as follows. The probability that a model will predict a pairwise consistent label for a pair of photographs is ${ {4\choose 2} = \frac{1}{2}}$. This is because there are four possible pairs of labels: (gay, gay), (gay, straight), (straight, straight) and (straight, gay). Two out of these four, (gay, gay) and (straight, straight), indicate that the model considers the subjects of the photographs to have the same sexual orientation.

Since we have three pairs of photographs, the probability of the model being pairwise consistent for all three pairs is: $$ \left(\frac{1}{2}\right)^3 = \frac{1}{8} $$
Then, if we consider the models to be independent, the probability of all three models being pairwise consistent is: $$ \left(\frac{1}{8}\right)^3 = \frac{1}{512} $$
\subsubsection*{Probability of consistent predictions across models}
Again assuming that the models are independent, the probability of all three models agreeing on the same sexual orientation label for a pair of photographs is $ \frac{2}{64} $. This is because the models have to predict either ((straight, straight), (straight, straight), (straight, straight)) or ((gay, gay), (gay, gay), (gay, gay)) out of $ 64 $ combinations. The probability of two out of three models agreeing on the label is $ \frac{6}{64} $. The remaining probability is $ \frac{56}{64} $. These probabilities are used to calculate how likely it is that two, three or no models agree on the sexual orientation label for the three pairs of photographs.
\section{Study 3: Test correlation of sexual orientation with head pose}
\label{study:study3_headpose}
To test whether the deep learning and facial morphology classifiers might be predicting sexual orientation from the \textit{head pose}, we test whether the three \textit{head pose} angles in the dataset are correlated with sexual orientation.

The \textit{Face++} service identifies three \textit{head pose} angles for each photograph: pitch, roll and yaw.

For each of these angles we measured the correlation between the angle and the binary category label (straight or gay) using a Point-Biserial Correlation \cite{pointbiserial}. We checked that the distribution for each angle is normally distributed using a Shapiro-Wilk test (the test was performed for each continuous variable in each binary category) \cite{shapiro}.

In addition, we investigated whether there might be a non-parametric relationship between the \textit{head pose} angles and sexual orientation. To do this we calculated the maximal information coefficient \cite{mic_reshef}.

All correlations were measured separately for males and females.
\section{Study 4: Test prediction of sexual orientation while controlling for facial hair and eyewear}
\label{study:study4_facial_hair_eyewear}
To test whether the ML models in Study 1 are learning to predict sexual orientation from acquired features such as the presence of facial hair or eyewear, the dataset was filtered so that these features occur with equal probability. Each photograph was manually labelled to indicate whether the subject had facial hair (\textbf{A}) or was wearing eyewear (\textbf{B}), as described in Section~\ref{section_dataset}.
\subsection*{A. Facial hair}
The ML experiments in Study 1 were repeated, but the dataset was filtered so that each photograph in each category has a 50\% chance of having facial hair or not. We only tested facial hair for male subjects. None of the female photographs exhibited significant facial hair.

We created a logistic regression model to predict sexual orientation from a cropped facial image in the same manner as in Study 1, the only difference is that the models were trained on this reduced dataset.

Table \ref{table:facial_hair_counts} shows the number of photographs in each category evenly balanced between those with and without facial hair.
\begin{table}[hb]
	\centering
	\begin{tabular}{lll}
		\hline
		\rowcolor[HTML]{000000} 
		{\color[HTML]{FFFFFF} } & {\color[HTML]{FFFFFF} Facial Hair} & {\color[HTML]{FFFFFF} No Facial Hair} \\ \hline
		Male - gay              & 1969                                & 1969                                   \\
		Male - straight         & 1969                                & 1969                                   \\ \hline
	\end{tabular}
	\caption{Number of photographs in each category with and without facial hair}%
	\label{table:facial_hair_counts}%
\end{table}
\subsection*{B. Eyewear}
A separate experiment was performed where the photographs in each category had a 50\% chance of having eyewear. Since such a small proportion of straight females have eyewear in their photographs (4\% versus 20\% for straight men), the sample size was too small to include females.

Table \ref{table:eyewear_counts} shows the number of photographs in each category evenly balanced between those with and without eyewear.

As with \textbf{A}), we used the same methodology to predict sexual orientation as in Study~1 but with the reduced dataset.
\begin{table}[hb]
\centering
	\begin{tabular}{lll}
		\hline
		\rowcolor[HTML]{000000} 
		{\color[HTML]{FFFFFF} } & {\color[HTML]{FFFFFF} Eyewear} & {\color[HTML]{FFFFFF} No Eyewear} \\ \hline
		Male - gay              & 965                                & 965                                   \\
		Male - straight         & 965                                & 965                                   \\%\\
		 \hline
	\end{tabular}
	\caption{Number of photographs in each category with and without eyewear}%
	\label{table:eyewear_counts}%
\end{table}
\section{Summary}
This chapter described the methodology for four experiments investigating whether it is possible to predict sexual orientation from facial images and which properties of these images might make that possible. The next section presents the obtained results.
\chapter{Results}
\label{chap:results}
\graphicspath{{figures/}}
\graphicspath{{chapters/introduction/figures/}}
This chapter presents some of the statistical properties of the final dataset (Section~\ref{results:dataset}), and the detailed results for each study from Chapter~\ref{chap:methods}. Section~\ref{results:study1} contains the results for each of the three ML models evaluated on the dataset of dating profile images. Section~\ref{results:study2} presents the results for the three ML models evaluated on photographs with intentionally altered presentation. Section~\ref{results:study3} contains the evaluation of whether the \textit{head pose} is correlated with sexual orientation. Section~\ref{results:study4} describes the results when predicting sexual orientation while controlling for facial hair and eyewear. Section~\ref{results:summary} briefly summarises the results of each experiment.
\section{Dataset}
\label{results:dataset}
Statistical properties of the final dataset are presented in this section.
\subsubsection*{Photograph count}
The final dataset used for the studies has 20,910 facial images of 10,372 gay and straight (55\%/46\%) men and women (49\%/51\%), see Table \ref{table:results_final_count} for details. Figure \ref{fig:figure_histogram_images_per_subject} shows the number of photographs per subject for males and females.
\subsubsection*{Skew by age}
\begin{figure*}[!tp]
	\centering
	\centerline{\includegraphics[width=6in]{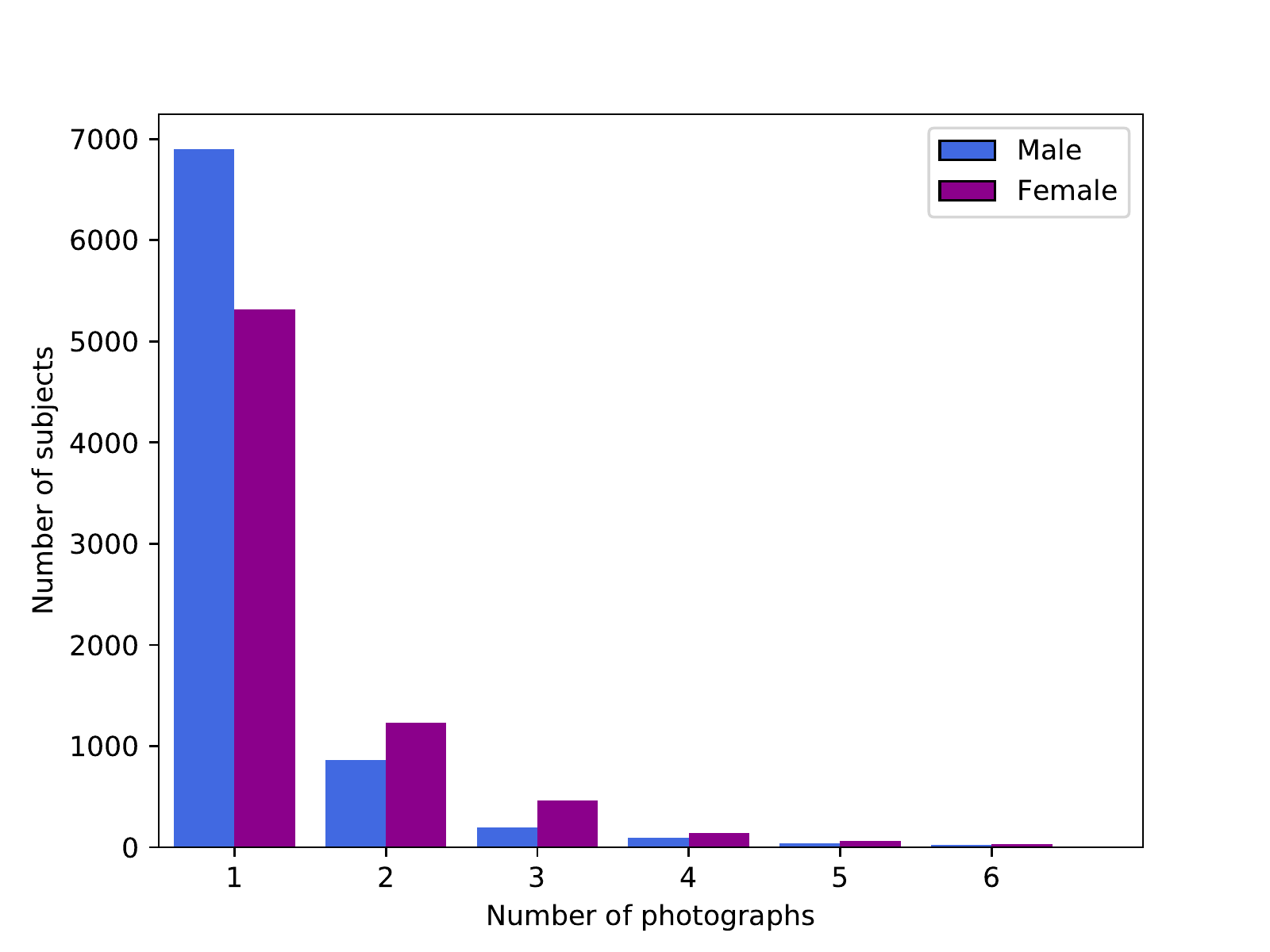}}%
	\caption{Number of photographs per subject}%
	\label{fig:figure_histogram_images_per_subject}%
	\vspace{0.2in}
	\centerline{\includegraphics[width=6in]{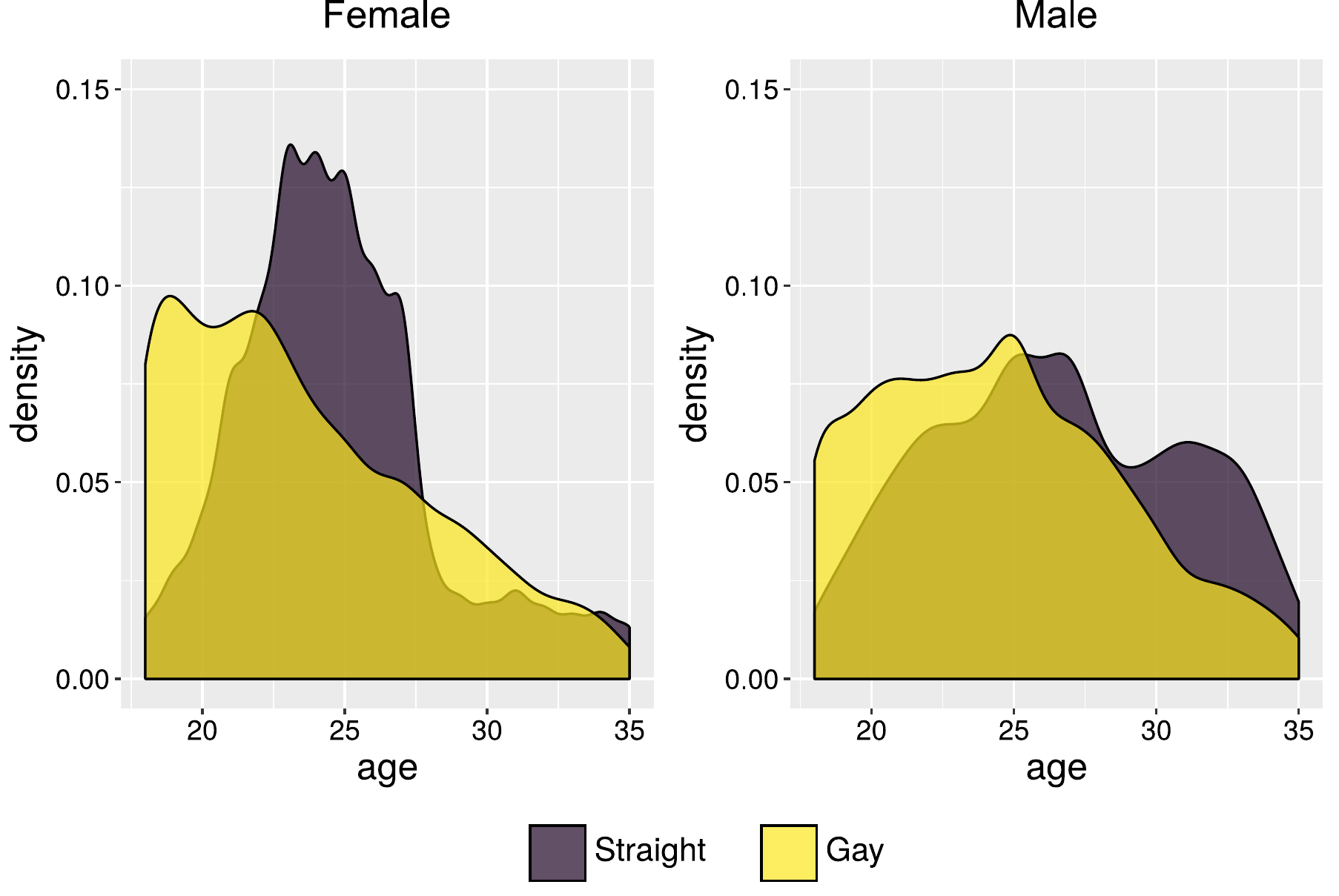}}%
	\caption{Age distribution of females and males by sexual orientation.}%
	\label{fig:age_density}%
\end{figure*}

The dataset has different age distributions for straight and gay subjects. Figure \ref{fig:age_density} shows that in the dataset gay females are younger than straight females. A similar skew exists for males.

The female age densities show steep peaks at around 20 years for gay subjects and~25 for straight subjects. The male distributions are more evenly spread out.
\subsubsection*{Composite facial images}
Appendix \ref{app:appendix1} shows composite photographs comparing averaged gay and straight faces from this dataset.
\begin{table}[!t]
	\centering
	\begin{tabular}{lcccc}
		\hline		
		& \multicolumn{2}{c}{Females} & \multicolumn{2}{c}{Males} \\ \cline{2-5} 
		& Gay        & Straight       & Gay       & Straight      \\ \hline
		Unique users         & 3760       & 3515           & 4738      & 3418          \\
		Median age (IQR)     & 24 (23--27) & 23 (20--27)     & 24 (21--28) & 26 (23--30) \\
		Total images         & 5132       & 5406           & 4666      & 5706          \\
		Users with at least: &            &                &           &             \\
		\,\,\,\,\,\,\,\,1 image              & 3760       & 3515           & 4738      & 3418          \\
		\,\,\,\,\,\,\,\,2 images             & 853        & 1103           & 652       & 604           \\
		\,\,\,\,\,\,\,\,3 images             & 270        & 458            & 147       & 246           \\
		\,\,\,\,\,\,\,\,4 images             & 92         & 173            & 72        & 128           \\
		\,\,\,\,\,\,\,\,5 images             & 45         & 76             & 36        & 70            \\ \hline
	\end{tabular}
	\caption{Frequencies of users and facial images and the median and interquartile range (IQR) for ages.}
	\label{table:results_final_count}
\end{table}
\clearpage
\section{Study 1: Prediction of sexual orientation from facial images using machine learning models}
\label{results:study1}
The results for the three ML models are presented.
\subsection{Model 1: Deep neural network classifier}
The classifier based on DNN features (Model~1) has a ROC AUC score of AUC=.68 for males and AUC=.77 for females when using one facial image to predict sexual orientation.

\begin{figure*}[t]
	\centering
	\centerline{\includegraphics[width=5.1in]{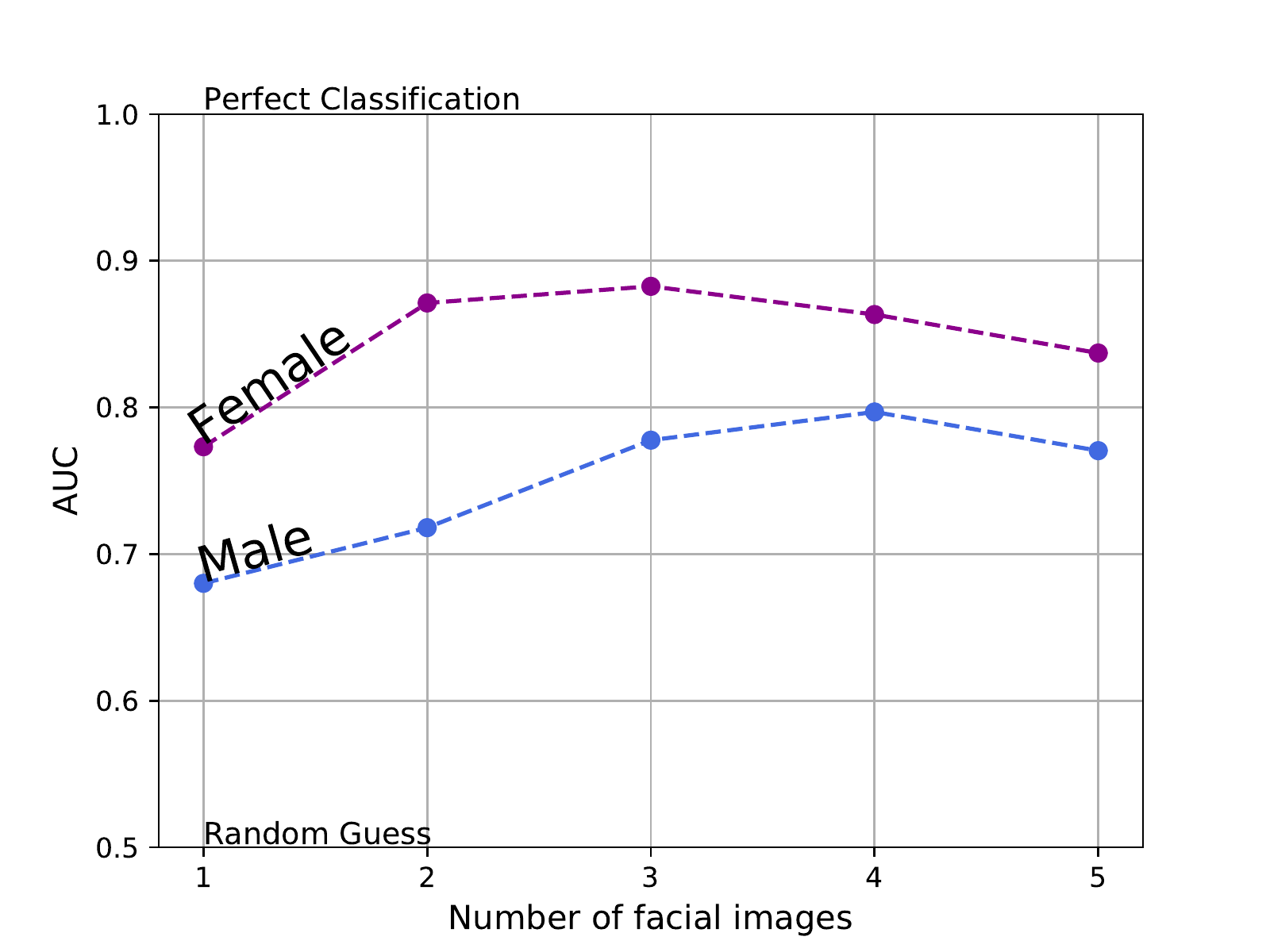}}%
	\caption{Accuracy of the DNN classifiers (Model~1) when provided with increasing numbers of images per subject.}%
	\label{fig:figure_1_vgg_auc_scores_by_num_images}%
\end{figure*}

Figure \ref{fig:figure_1_vgg_auc_scores_by_num_images} shows the AUC scores for Model~1 with increasing numbers (from 1 to 5) of images per subject. The accuracy with which it predicts sexual orientation increases with more images per subject. At three images per subject the classifier scores AUC=.78 for males and AUC=.88 for females.

The accuracy decreases with more than three images. This is due to the very small sample size available for subjects with four and five images available in the dataset (see~Table~\ref{table:results_final_count}). The performance values for four and five subjects also vary significantly every time that the classifier is scored, due to the random sample taken during the scoring procedure (see Section~\ref{section:model_scoring}). For example, when evaluating four images per subject the minimum and maximum scores recorded for female subjects were AUC=.84 and AUC=.87, respectively (s.d. 0.01). For males with four images the minimum and maximum were AUC=.78 and AUC=.83, respectively (s.d. 0.01). For the remaining studies only scores up to three images per subject are reported.

\begin{figure*}[!t]
	\centering
	\centerline{\includegraphics[width=6in]{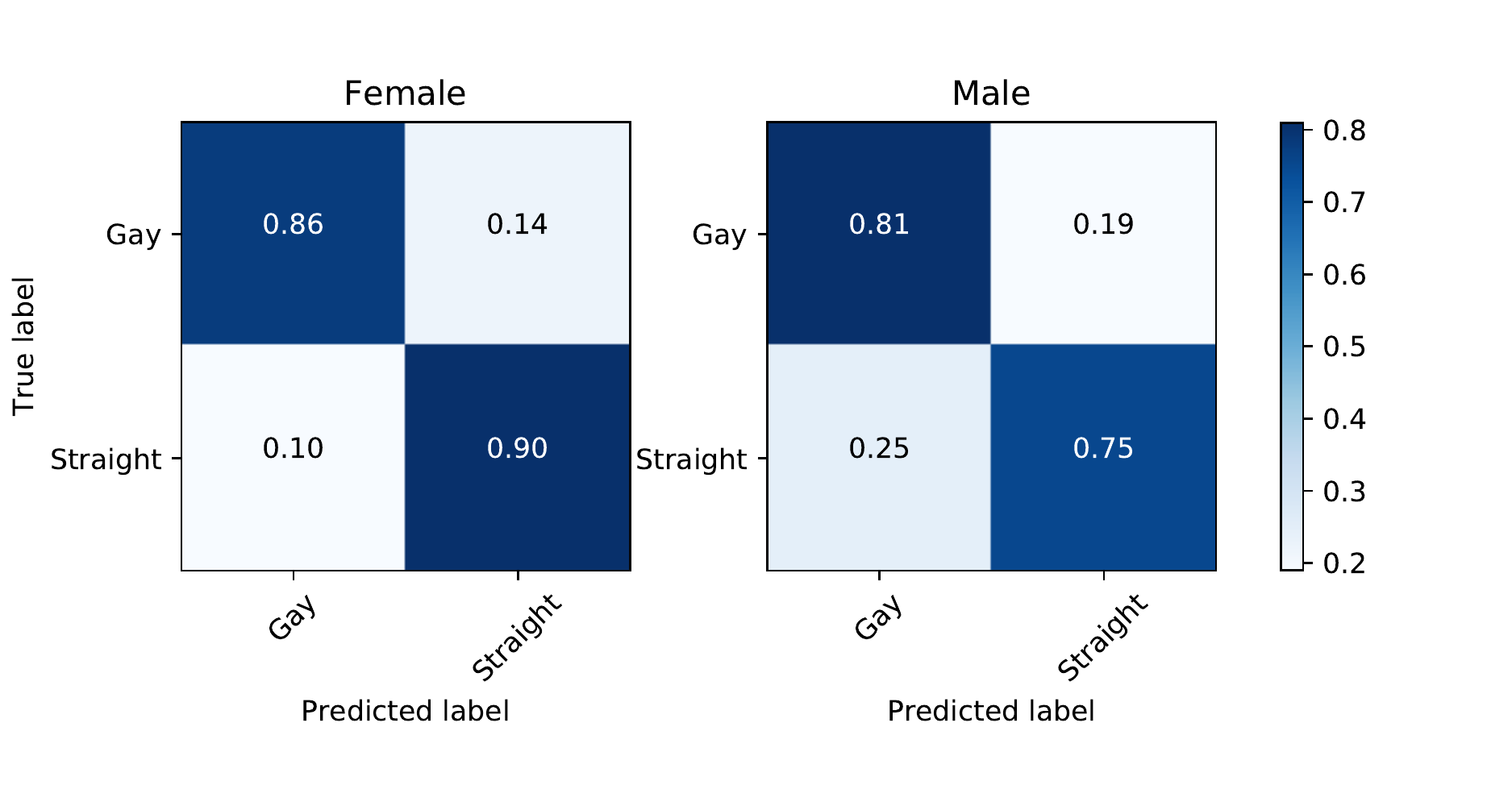}}%	
	\caption{Confusion matrices for the DNN classifier (Model~1) when provided with three facial images per subject.}%
	\label{fig:study_1_confusion_matrix}%
\end{figure*}

Figure \ref{fig:study_1_confusion_matrix} shows the confusion matrices\footnote{A confusion matrix shows the proportion of true positives, false positives, true negatives and false negatives in four quadrants.} for Model~1's female and male classifiers predicting sexual orientation from three images per subject.
\subsection{Model 2: Facial morphology classifier}
\label{results:study2}
\begin{figure*}[!htbp]
	\centering
	\centerline{\includegraphics{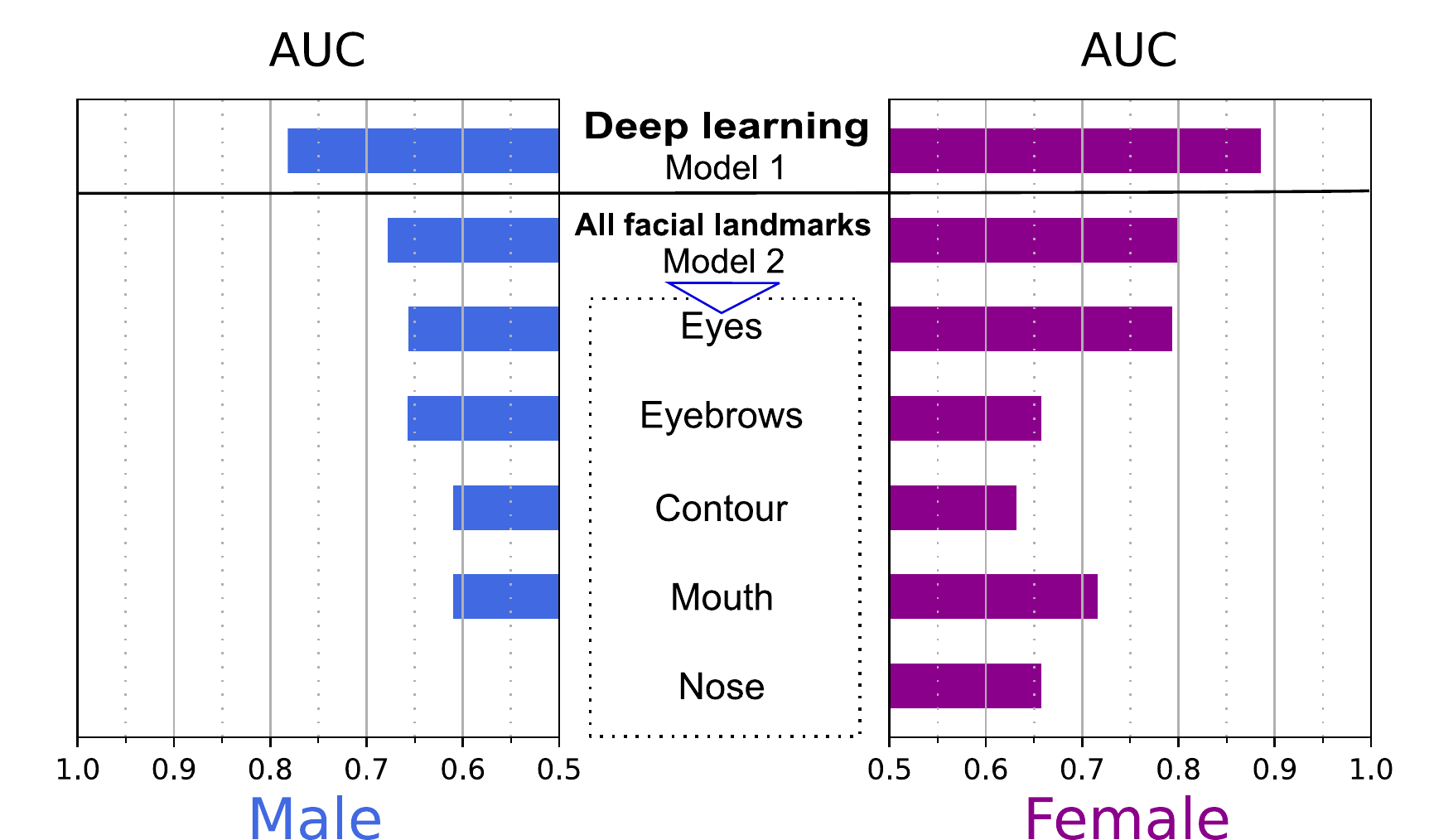}}
	\caption{Accuracy of the facial morphology classifiers (Model~2) when provided with three images per subject. For comparison the accuracy of the deep learning classifier, Model~1, is shown on top (for three images per subject).}
	\label{fig:figure_2_vgg_and_fpp_scores_by_component}%
\end{figure*}
The classifier based on facial morphology (Model~2) has a ROC AUC score of AUC=.62 for males and AUC=.72 for females when predicting sexual orientation from the whole face using one facial image per subject. When provided with three images per subject, Model~2 scored AUC=.68 for males and AUC=.81 for females. 

Figure \ref{fig:figure_2_vgg_and_fpp_scores_by_component} shows the AUC scores for each facial component when provided with three images per subject. The results from Model~1 are shown on top for reference.

For males the eyes and eyebrows are most predictive of sexual orientation and the nose has no predictive value. For females the eyes are most predictive and the facial contour least predictive.

As with Model~1 for this dataset, the results for females are significantly higher than for males.
\subsubsection*{Limitations}
The facial morphology features used by this model may be sensitive to \textit{head pose} \cite{stereotypes}. They may also be sensitive to smiling or the overall facial expression.
\subsection{Model 3: Highly blurred image classifier}
\label{results:study3}
The classifier based on highly blurred images (Model~3) has a ROC AUC score of AUC=.63 for males and AUC=.72 for females when predicting sexual orientation from one 5$\plh$5 pixel image. When provided with three images per subject, this increases to male AUC=.63 and female AUC=.82.

\begin{figure*}[!htbp]
	\centering
	\centerline{\includegraphics{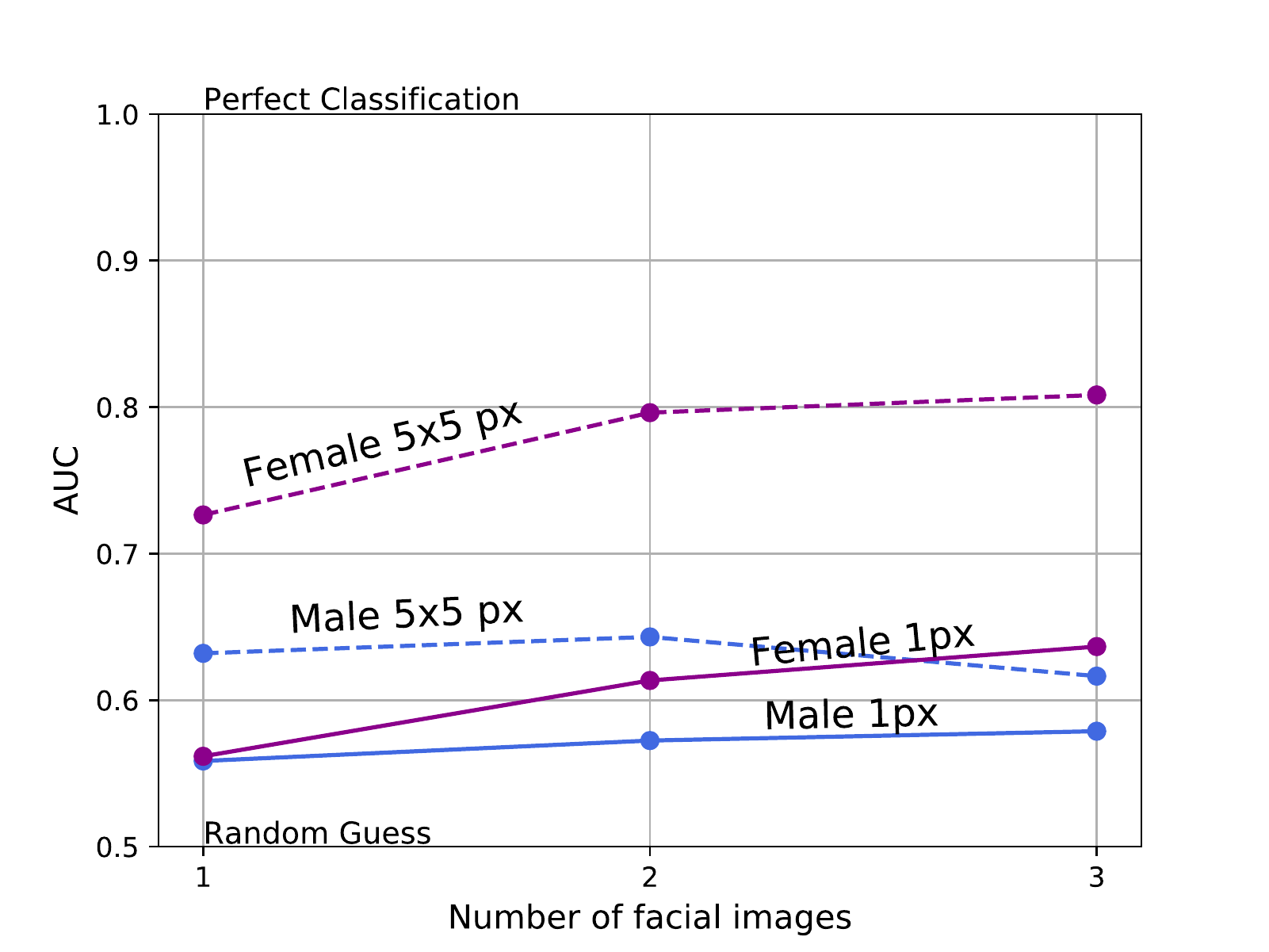}}%
	\caption{Accuracy of the classifier derived from Highly Blurred Images (Model~3). Scores are shown with increasing numbers of images per subject.}
	\label{fig:figure_blurred_1_scores_by_num_images}%
\end{figure*}

Figure \ref{fig:figure_blurred_1_scores_by_num_images} shows the AUC scores for Model~3 with increasing numbers (from 1 to 3) of images per subject. It shows results for both 5$\plh$5 pixel images and 1 pixel images.

The classifier derived from 5$\plh$5 pixel images shows accuracy comparable with the facial morphology classifier (Model~2).
\section{Study 2: Evaluate machine learning models with altered presentation}
\label{results:study2}
\begin{figure*}[!htbp]
	\centering
	\centerline{\includegraphics[width=4.5in]{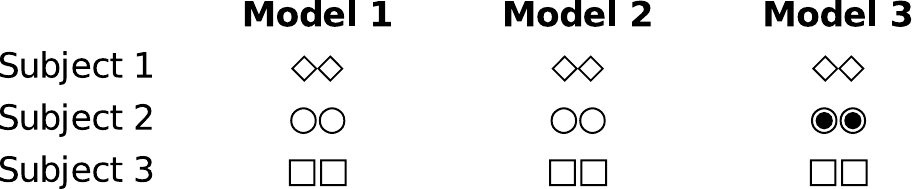}}%
	\caption{Classification results for each facial image. Original photographs are shown in Figures \ref{fig:figure_altered_presentation}. Each model was used to predict sexual orientation for the pair of photographs for each subject. Matched symbols for a subject indicate a pairwise consistent prediction by that model. Matching symbols per subject across models indicate agreement on sexual orientation by the models for that subject.}%
	\label{fig:figure_three_subjects_three_models}%
\end{figure*}

All three ML models were evaluated on pairs of photographs with intentionally altered presentation. The predicted sexual orientation labels are show in Figure \ref{fig:figure_three_subjects_three_models} using symbols to depict matching labels. All three models predicted pairwise consistent labels for the three pairs of photographs in Figure \ref{fig:figure_altered_presentation}.

\subsubsection*{Probability of pairwise consistency}
Assuming that the models are independent and unbiased there is a
\begin{equation}
\left(\frac{1}{8}\right)^3 = \frac{1}{512} = 0.2\%
\end{equation}
chance of all three models predicting a pairwise consistent label for all pairs of photographs.
%\end{samepage}
\subsubsection*{Probability of consistent predictions across models}
The sexual orientation labels for each subject were also predicted consistently across all the models (with the exception of Model~3 for Subject~2).

%\begin{samepage}
With the same assumptions, there is a
\begin{equation}
	\frac{2}{64} \times \frac{2}{64} \times \frac{6}{64} = 0.01\%
\end{equation}
chance of all three models agreeing on the same pairs of sexual orientation labels for two subjects and two models agreeing for the other subject.
%\end{samepage}

\subsection{Limitations}
This experiment had a very small sample size of only six facial images. The Highly Blurred Image classifier (Model~3) could possibly benefit from the fact that these photographs are taken under similar conditions with similar lighting, similar backgrounds and the same colour of clothing.
\section{Study 3: Test correlation of sexual orientation with head pose}
\label{results:study3}
No strong correlations were found between any of the \textit{head pose} angles and the sexual orientation of the subjects.
\begin{table}[!htbp]
\centering
	\begin{tabular}{ccccc}
		\cline{1-3}
		\rowcolor[HTML]{000000} 
		\multicolumn{1}{|c|}{\cellcolor[HTML]{000000}{\color[HTML]{FFFFFF} \textbf{Angle type}}} & \multicolumn{2}{c|}{\cellcolor[HTML]{000000}{\color[HTML]{FFFFFF} \textbf{Correlation}}} & \multicolumn{2}{c}{\cellcolor[HTML]{000000}{\color[HTML]{FFFFFF} \textbf{MIC}}} \\ \cline{1-3}
		\rowcolor[HTML]{000000} 
		{\color[HTML]{FFFFFF} }                                                                  & {\color[HTML]{FFFFFF} Female}                & {\color[HTML]{FFFFFF} Male}               & {\color[HTML]{FFFFFF} Female}           & {\color[HTML]{FFFFFF} Male}           \\
		\multicolumn{1}{c|}{Pitch}                                                               & \multicolumn{1}{c|}{0.10}                    & \multicolumn{1}{c|}{0.03}                 & \multicolumn{1}{c|}{0.07}               & 0.06                                  \\
		\multicolumn{1}{c|}{Roll}                                                                & \multicolumn{1}{c|}{0.05}                    & \multicolumn{1}{c|}{0.16}                 & \multicolumn{1}{c|}{0.07}               & 0.08                                  \\
		\multicolumn{1}{c|}{Yaw}                                                                 & \multicolumn{1}{c|}{0.03}                    & \multicolumn{1}{c|}{-0.01}                & \multicolumn{1}{c|}{0.06}               & 0.06                                 
	\end{tabular}
	\caption{Correlation and maximal information coefficient (MIC) relative to sexual orientation for each \textit{head pose} angle.}
	\label{table:angle_correlations}
\end{table}

Table \ref{table:angle_correlations} lists the Point-Biserial correlation \cite{pointbiserial} and maximal information coefficient~\cite{mic_reshef} for each type of \textit{head pose} angle (pitch, roll and yaw) with the sexual orientation of the subject.

All correlations had a p-value of less than 0.05, except for the uncorrelated yaw angle for males.

\begin{figure*}[!htbp]
	\centering
	%	\centerline{\includegraphics{angle_density_female}}
	%	\centerline{\includegraphics{angle_density_male}}
	\centerline{\includegraphics{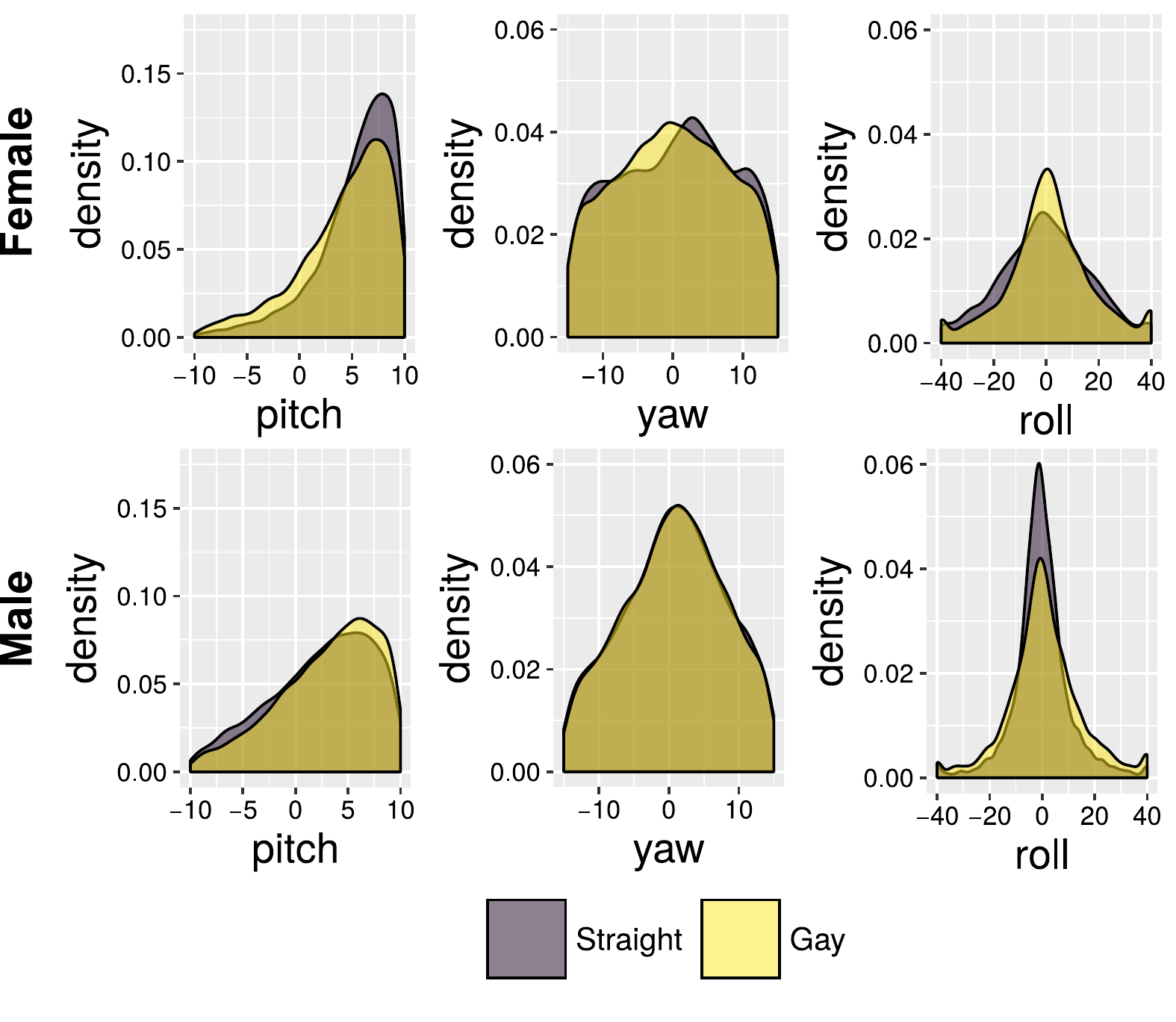}}
	\caption{Distribution of \textbf{pitch}, \textbf{yaw} and \textbf{roll} angles by sexual orientation.}
	\label{fig:figure_angle_categories}%
\end{figure*}

Although the correlations are weak, it is possible to see differences related to sexual orientation in the distribution plots. Figure \ref{fig:figure_angle_categories} plots the distributions of the pitch, roll and yaw angles of the \textit{head pose} for gay and straight subjects\footnote{Only photographs with pitch angles of less than 10 degrees and yaw angles less than 15 degrees were included in the dataset. The roll angles were clipped at -40 degrees and 40 degrees for these plots.}.

Although both categories of females favour photographs taken from a higher angle (greater pitch), it is more pronounced in straight females. Straight females are also more likely to tilt (roll) their heads. There also appears to be a small difference in preference for turning the chin to one side (yaw).

Gay males have a slight preference for a greater pitch and are more likely to tilt their heads (roll).
\subsection*{Limitations}
The correlation test only evaluated each angle individually against the sexual orientation category. It could be that all three angles together show a greater correlation with sexual orientation.

The \textit{head pose} information is provided by the \textit{Face++} model and it is assumed that this information is accurate.
\section{Study 4: Test prediction of sexual orientation while controlling for facial hair and eyewear}
\label{results:study4}
\begin{figure*}[!h]
	\centering
	\centerline{\includegraphics{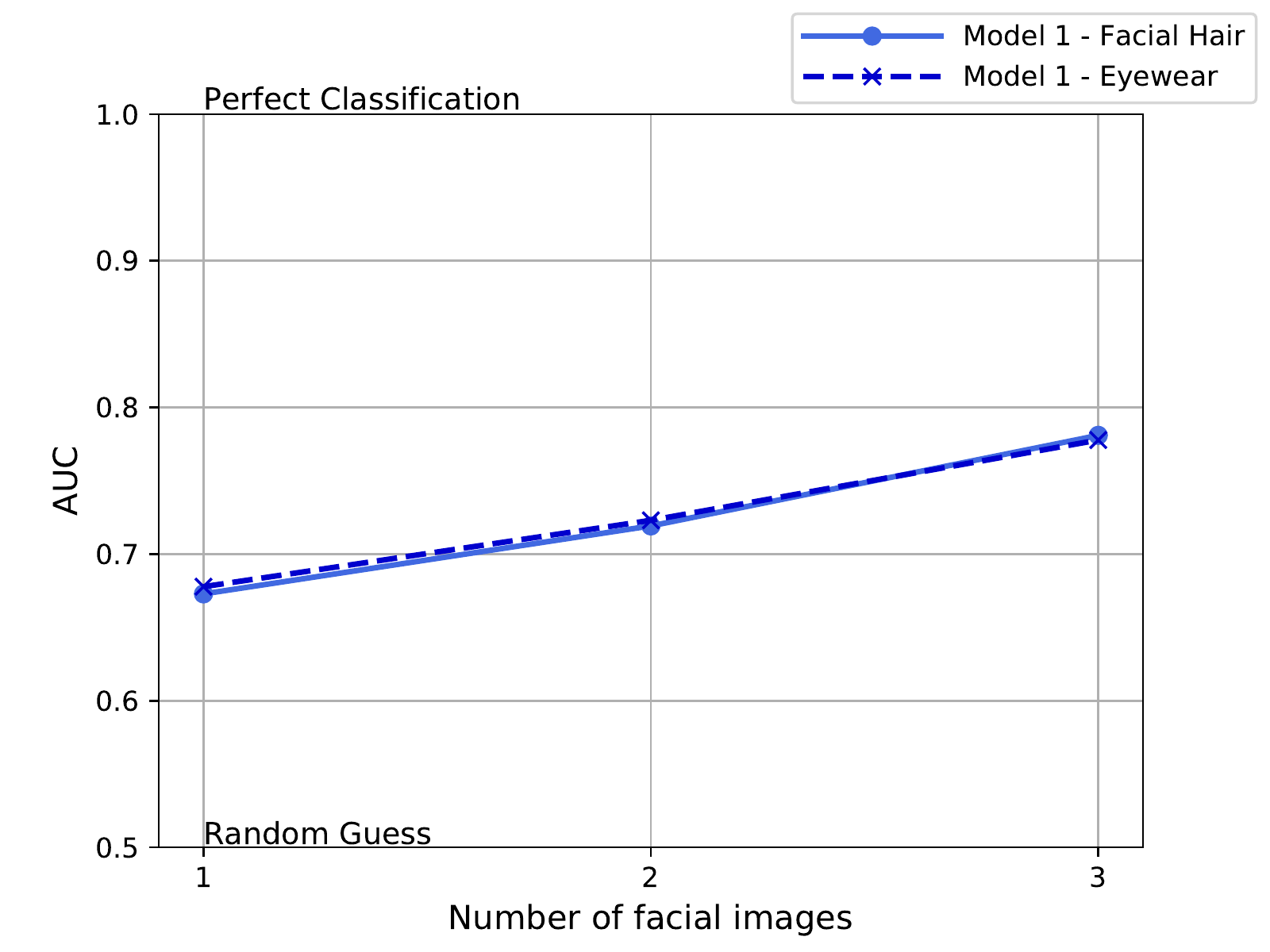}}
	\caption{Accuracy in predicting sexual orientation for males evenly split between those with and without \textbf{facial hair} and \textbf{eyewear}. Scores were generated using Model~1 and are shown for an increasing number of images available per subject.}%
	\label{fig:figure_all_model_scores_facial_hair}%
\end{figure*}

Figure \ref{fig:figure_all_model_scores_facial_hair} shows that when gay male subjects and straight male subjects are evenly split between those with and those without \textbf{facial hair}, the classifier is still able to predict sexual orientation accurately (male AUC=.67; for three images per subject AUC=.77).

The same applies when the subjects are evenly split between those with and without \textbf{eyewear} in each category (male AUC=.67; for three images per subject AUC=.78).
\clearpage
\section{Summary of results}
\label{results:summary}
A summary of the results for each study is presented.
\subsubsection*{Study 1}
The results for Study 1 show that ML models are capable of predicting sexual orientation from facial images, specifically, photographs from online dating profiles. The two ML models (Model~\hyperref[model:vgg]{1} and Model~\hyperref[model:fpp]{2}) replicating experiments by W\&K \cite{wang_kosinski} using a new dataset show that it is possible to predict sexual orientation from both photographs (see~Figure~\ref{fig:figure_fpp_crop}) and facial morphology (see Figure \ref{fig:figure_fpp_example}).

In addition, the new blurred image classifier (Model~\hyperref[model:blurred]{3}) demonstrates that the colour information contained in a blurred photograph of the face (see Figure \ref{fig:figure_study_3_blurred_image}) is also predictive of sexual orientation.

\begin{figure*}[!ht]
	\centering
	\centerline{\includegraphics{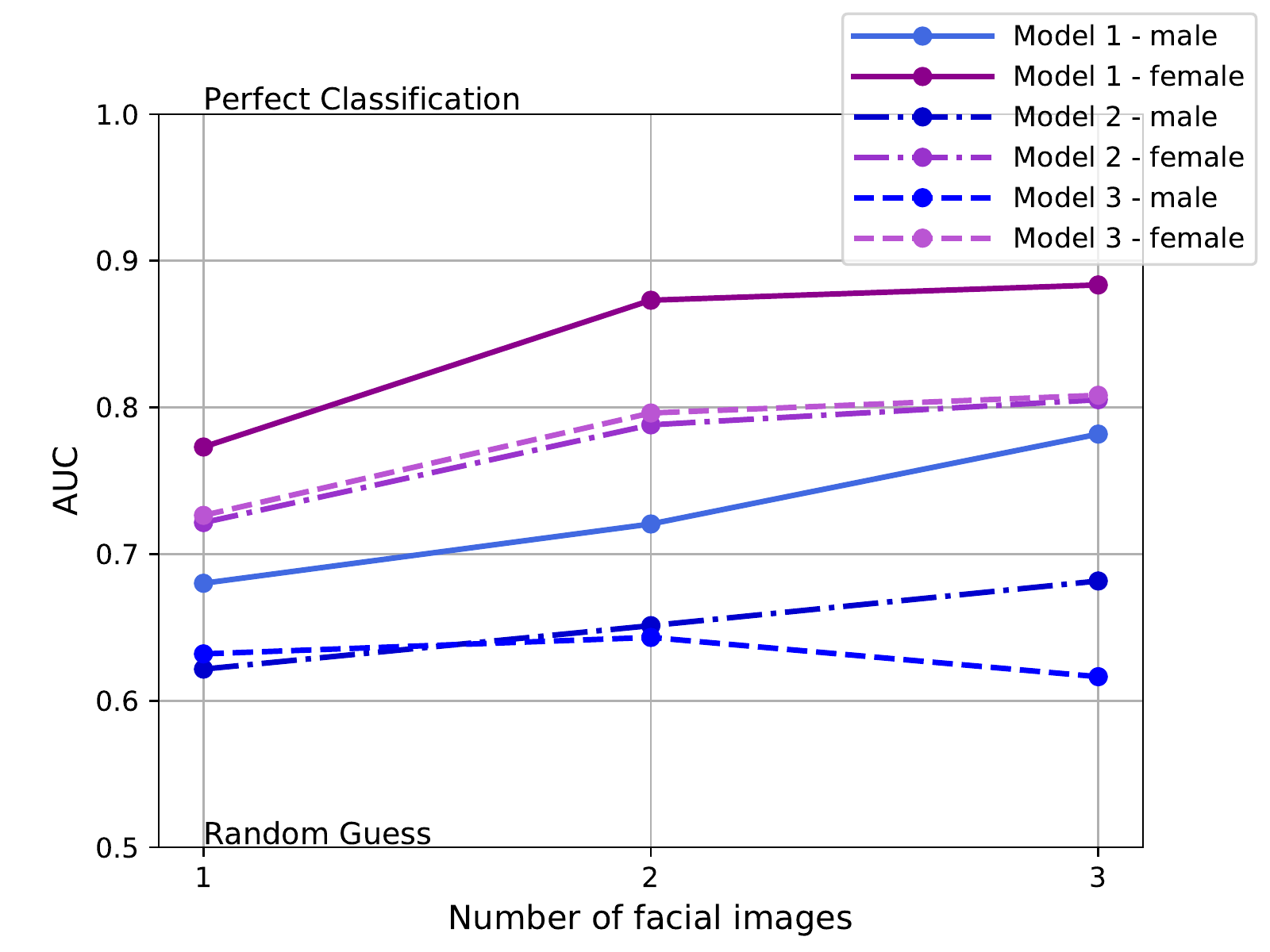}}%
	\caption{Accuracy predicting sexual orientation for Models 1, 2 and 3. Scores are shown for increasing number of images per subject.}%
	\label{fig:figure_all_model_scores}%
\end{figure*}

Figure \ref{fig:figure_all_model_scores} compares the performance for all three models, for both male and female datasets. ROC AUC scores are shown for increasing numbers of photographs per subject.% 
\subsubsection*{Study 2}
The models above were tested on pairs of photographs in which individuals modified their appearance (see Figure \ref{fig:figure_altered_presentation}) to appear more stereotypically gay or straight. Changes to presentation included large changes in pitch and changes to facial hair, eyewear and makeup. All three classifiers were pairwise consistent in predicting the same sexual orientation label for both photographs in each pair.
\subsubsection*{Study 3}
The angles of the head in the photograph making up the \textit{head pose} (pitch, roll and yaw) are not correlated with the subject's sexual orientation. 
\subsubsection*{Study 4}
When male subjects are evenly split in each category between those with facial hair and those without facial hair, the classifier is still able to predict sexual orientation accurately (male AUC=.68). Similarly for eyewear, when male subjects are evenly split between those with and those without, the classifier is still able to predict sexual orientation accurately (male AUC=.66).
\subsubsection*{Summary}
\label{sec:intro1}
\addcontentsline{toc}{subsection}{\nameref{sec:intro1}}
This chapter presented the results for the four experiments described in Chapter \ref{chap:methods}. Chapter \ref{chap:discussion} discusses how the results of the replication study compare with that of the original study and what the results of the other experiments reveal about the ability to predict sexual orientation.
\chapter{Discussion}
\label{chap:discussion}
This chapter discusses the results obtained for studies 1-4.
\subsubsection*{Replication and comparison with previous work}
The previous experiments by W\&K replicated here show that ML techniques can predict sexual orientation from facial images. They can also do so better than the human judges tested in their control experiment (human judges scored male AUC=.61 and female AUC=.54 in their experiment \cite{wang_kosinski}).

Despite the smaller dataset (this study has about 20,000 images, where W\&K used 35,000), the models in this study have broadly similar accuracy. However, the DNN model in this study is more accurate for females than for males (Model~1 male AUC=.78 and female AUC=.88 for three images per subject). W\&K's model performs better for males than females (W\&K male AUC=.88 and female AUC=.76 for three images per subject).

Similarly for facial morphology, the classifier in this study performs better for females than for males (Model~2 male AUC=.68 and female AUC=.81 for three images per subject). W\&K's model performs better for males than females (W\&K male AUC=.85 and female AUC=.70 for five images per subject).

Each facial component in this study (except for noses for males) was also predictive of sexual orientation, but the relative ranking compared to W\&K's results was completely different. In W\&K's study, the facial contour component was as good or better than all the other components. In this study, the eyes were the best predictor, and the contour was as poor as or was the least predictive component.
\subsubsection*{Differences in datasets}
While W\&K limited their study to white subjects from the United States, this study included facial images from individuals from many different countries, representing several broad racial groups. Most of the images in this study are of whites and asians. The subjects are younger than those in W\&K (see Table \ref{table:results_final_count}).

A significant difference between this dataset and W\&K's is that the data used in this study has some number of straight transgender females. It is possible that differences in either their facial morphology or presentation (or both) make it easier for the female classifiers to distinguish gay facial images from straight images. This could explain why this study's ML models all perform better for the female dataset than for males.

When comparing the composites in Appendix \ref{app:appendix1} against the composites published by W\&K \cite{wang_kosinski}, the ``baseball cap'' effect of a darkened forehead is noticeably absent in this study's composites.
\subsubsection*{Differences in presentation versus biological differences}
It is difficult to determine whether the ML models employed in this study are predicting sexual orientation from differences that are of a biological nature (such as the shape of the face and facial features) or differences in presentation.

The results from Study 2 indicate that a deliberate change in presentation is not sufficient to change the sexual orientation prediction label for the three models used in this study. Furthermore, all the models (with one exception) agreed on the sexual orientation label for the three individuals tested.

Study 3 also found no correlation between \textit{head pose} angles and sexual orientation, implying that the classifiers are relying on additional information present in each facial image.

The fact that a model is able to predict sexual orientation from facial morphology (Model~2) implies either a biological difference, or an apparent difference due to \textit{head pose}, or some consistent difference in facial expression (such as smiling). Straight females in Appendix \ref{app:appendix1} appear to have a slightly more pronounced smile than their gay counterparts. Gay males in the composite images have clearly more pronounced smiles than their straight counterparts.

In the composite images it can be seen that the gay males have slightly thinner faces (and smaller jaws) than the straight males. Gay females have slightly wider faces and jaws than the straight females. These differences may have a biological origin, but they are also consistent with the pitch angle differences plotted in Figure \ref{fig:figure_angle_categories}.
\subsubsection*{Blurred photographs}
Model 3 showed that a model trained on a highly blurred facial image (Figure \ref{fig:figure_study_3_blurred_image}) is able to predict sexual orientation. The 1 pixel based classifier performed relative poorly, achieving scores barely better than chance for one subject per image. However, the 5$\plh$5~pixel model achieved a score of AUC=.63 for males and AUC=.72 for females using a single image per subject.

This indicates that there is a significant amount of information in the brightness, hue or other colour related information in the blurred images from this dataset that is predictive of sexual orientation. Inspection of the composites in Appendix \ref{app:appendix1} show noticeably brighter faces for straight females and gay males. Straight females also appear to wear brighter, pinker makeup on their lips.

Appendix \ref{app:appendix2} contains plots for the hue, saturation and brightness \cite{foley} distributions of the blurred images, and compares gay and straight sexual orientations. There is a clear difference in the brightness distributions for both females and males. The saturation values are also different for females.

Although the classifier is able to learn to predict sexual orientation from blurred photographs, the underlying cause for such differences is unknown. They could be due to a biological difference such as a difference in facial brightness. It could also be due to a group preference for makeup, or perhaps related to how the photograph is taken. Some people might use a mobile phone to take photographs for dating profiles and others might have them taken in a professionally lit photographic studio. The types of post-processing applied to photographs might vary between groups \cite{gelman}. It is also possible that photographs from different types of mobile phones or those that are uploaded to different dating websites are processed with different image compression algorithms and that there are artifacts resulting from these methods that are easily detectable by ML models.  
\subsubsection*{Controls for facial hair and eyewear}
The dataset used in this study confirms some of the trends regarding facial hair and eyewear investigated by Ag{\"u}era y Arcas {\it et al} \cite{stereotypes}. In the composite images more signs of facial hair are visible for straight white males as compared to gay white males. In the dataset for this study straight males are more likely to have facial hair (57\% versus 44\% for gay males). It is also possible to observe traces of spectacles on the gay males and females. In this dataset gay and straight males are equally likely to wear eyewear (20\%). Straight females are highly unlikely to wear eyewear (4\%) versus gay females (20\%).

Despite this preference for facial hair in straight males, Study 4 demonstrates that the models are still able to predict sexual orientation even when the proportion of subjects with facial hair in each category has been balanced out. The same applies for eyewear. This implies that neither differences in facial hair or eyewear alone are responsible for the ability to detect sexual orientation.
\chapter{Conclusion}
\label{chap:conclusion}
By means of a replication study, this work set out to investigate whether it is possible to predict sexual orientation from facial images. The replication results for two of W\&K's ML models verified the ability to predict sexual orientation from dating profile photographs. While the dataset used in this study produces better results for females than males, the accuracies of the models are broadly similar to that reported by W\&K~\cite{wang_kosinski}.

This study also demonstrated that a new ML model using highly blurred facial images is capable of predicting sexual orientation. This model relies on consistent differences in the colour information (such as hue, saturation and brightness) in the facial images to be able to distinguish between gay and straight subjects.

By testing the models on portraits intentionally altered with changes to makeup, facial hair, eyewear and the \textit{head pose}, it has been shown that the models are invariant to these changes. Furthermore, they are still capable of predicting sexual orientation while controlling for facial hair or eyewear.

These results leave open the question of how much the prediction of sexual orientation is influenced by biological features such as facial morphology, and how much by differences in presentation, grooming and lifestyle. Future work on this topic might investigate more precisely what role facial morphology has in predicting sexual orientation. It would also be useful to test whether ML models learn to predict sexual orientation from makeup. The nature of the relationship between colour information (such as facial brightness) with sexual orientation is not clear, future work might explore this.

%\include{chapters/chapter1/main} % rename as required (but remember to update directories appropriately)
%\include{chapters/chapter2/main} % rename as required (but remember to update directories appropriately)
% add all other chapters you decide to write (also create an appropriate directory structure)
%\include{chapters/conclusions/main}

%%%%%%%%%%%%%%%%%%%%%%%%%%%%%%%%%%%%%%%%%%%%%%%%%
%%%%%%%%%%%%%%%%%%%%%%%%%%%%%%%%%%%%%%%%%%%%%%%%%

\cleardoublepage
\ifpdf
\phantomsection
\fi
\label{bibliography}
\addcontentsline{toc}{chapter}{Bibliography}
%\bibliographystyle{plain}
%\bibliography{thesis}

%%%%%%%%%%%%%%%%%%%%%%%%%%%%%%%%%%%%%%%%%%%%%%%%%
%%%%%%%%%%%%%%%%%%%%%%%%%%%%%%%%%%%%%%%%%%%%%%%%%

\appendix
%%%%%%%%%%%%%%%%%%%%%%%%%%%%%%%%%%%%%%%%%%%%%%%%%
%%%%%%%%%%%%%%%%%%%%%%%%%%%%%%%%%%%%%%%%%%%%%%%%%

\chapter{Composite Photographs}
\label{app:appendix1}
\begin{figure*}[!htbp]
	\centering
	\centerline{\includegraphics[width=6.5in]{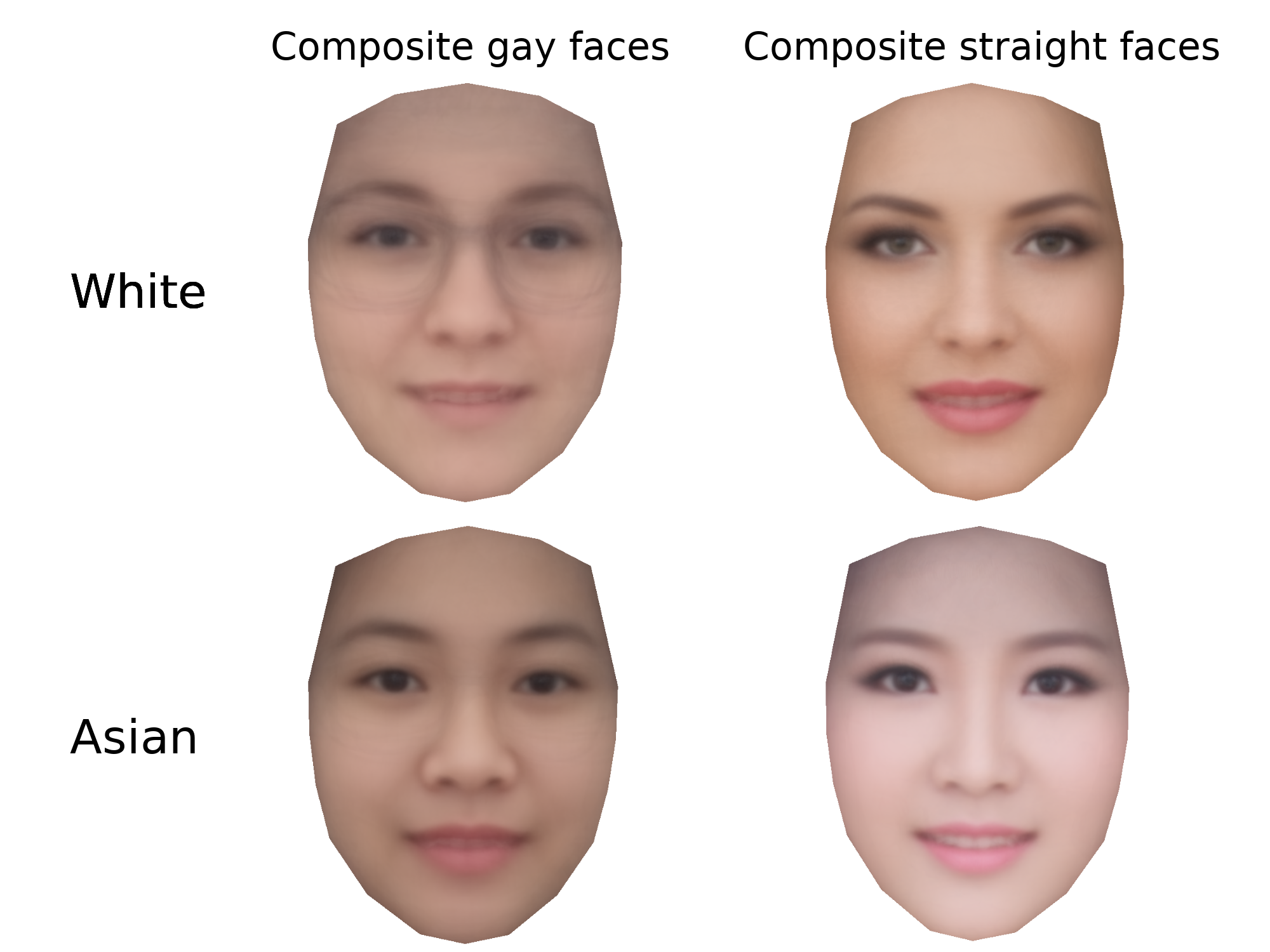}}
	\caption{Composite gay and straight female faces.}
	\label{fig:composite_female}%
\end{figure*}	
\begin{figure*}[!htbp]
	\centering
	\centerline{\includegraphics[width=6.5in]{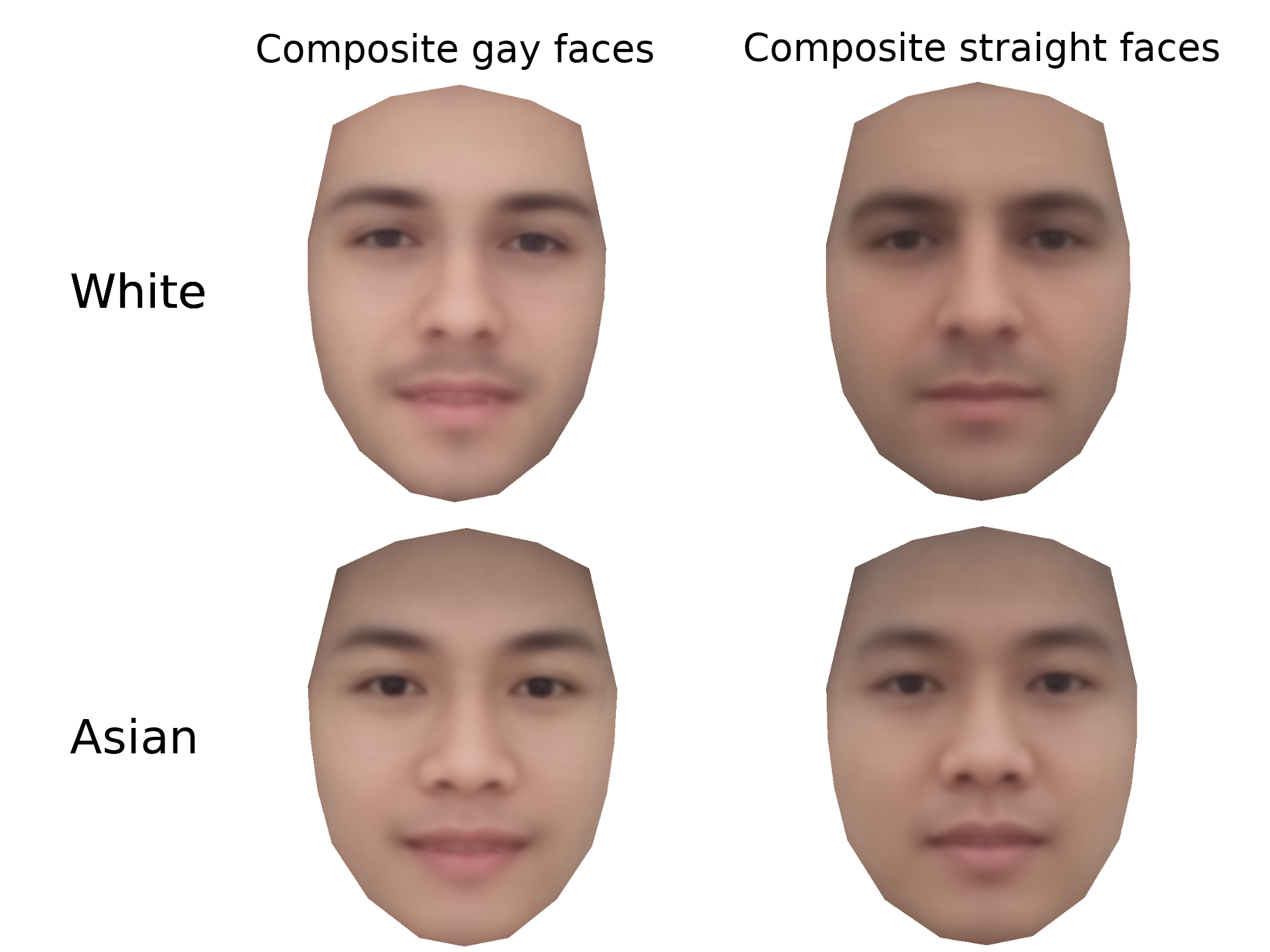}}
	\caption{Composite gay and straight male faces.}
	\label{fig:composite_male}%
\end{figure*}
Figures \ref{fig:composite_female} and \ref{fig:composite_male} show composite facial images of gay and straight subjects (Composites are shown separately for asian and white subjects).

To create the composites, images were first classified into broad racial groups. Composite images were only created for well represented groups (white and asian subjects). For each group the photographs were ranked by the ML Model~1 (deep learning classifier) from Study 1. The ranking orders photographs from those ``least likely to be gay'' (straight) to ``most likely to be gay''. Then photos were filtered to have a yaw angle of less than 11 degrees. Finally the top 150 ranked and bottom 150 ranked photographs were composed together to make up composite images \footnote{The compositor failed to recognize faces in some images but at least 100 images were used for every composite photograph}.

Photographs were composed by first locating the position of the main facial features in each photograph and then averaging them to create a target face. Then Delaunay triangulation was used to map each source image to the destination image. Each triangle in each source image was copied to the destination triangle using affine transformation and bilinear interpolation. Finally the pixels were averaged to obtain the composite facial image.
%%%%%%%%%%%%%%%%%%%%%%%%%%%%%%%%%%%%%%%%%%%%%%%%%
%%%%%%%%%%%%%%%%%%%%%%%%%%%%%%%%%%%%%%%%%%%%%%%%%

%\section{Summary}
%\label{sec:appendix1:summary}

%As always, provide a summary at the end.

%%%%%%%%%%%%%%%%%%%%%%%%%%%%%%%%%%%%%%%%%%%%%%%%%
%%%%%%%%%%%%%%%%%%%%%%%%%%%%%%%%%%%%%%%%%%%%%%%%%
%%%%%%%%%%%%%%%%%%%%%%%%%%%%%%%%%%%%%%%%%%%%%%%%%
%%%%%%%%%%%%%%%%%%%%%%%%%%%%%%%%%%%%%%%%%%%%%%%%%
\graphicspath{{appendices/appendix2/}}
\chapter{Hue, Saturation and Brightness}% of blurred images}
\begin{figure*}[!htbp]
	\centering
	\centerline{\includegraphics[width=6.2in]{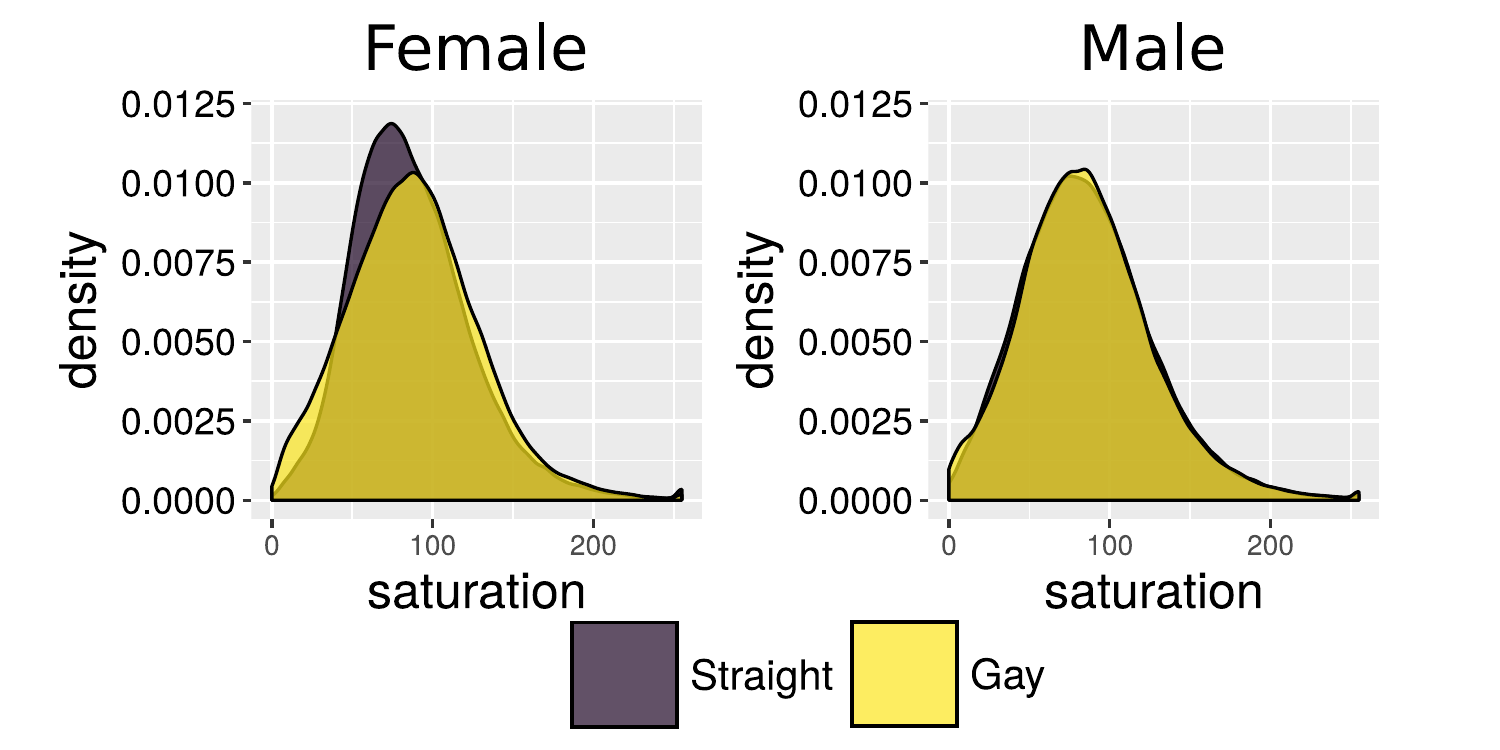}}
	\caption{Saturation}%
	\label{fig:saturation_density}%
\end{figure*}

\begin{figure*}[!htbp]
	\centering
	\centerline{\includegraphics[width=6.2in]{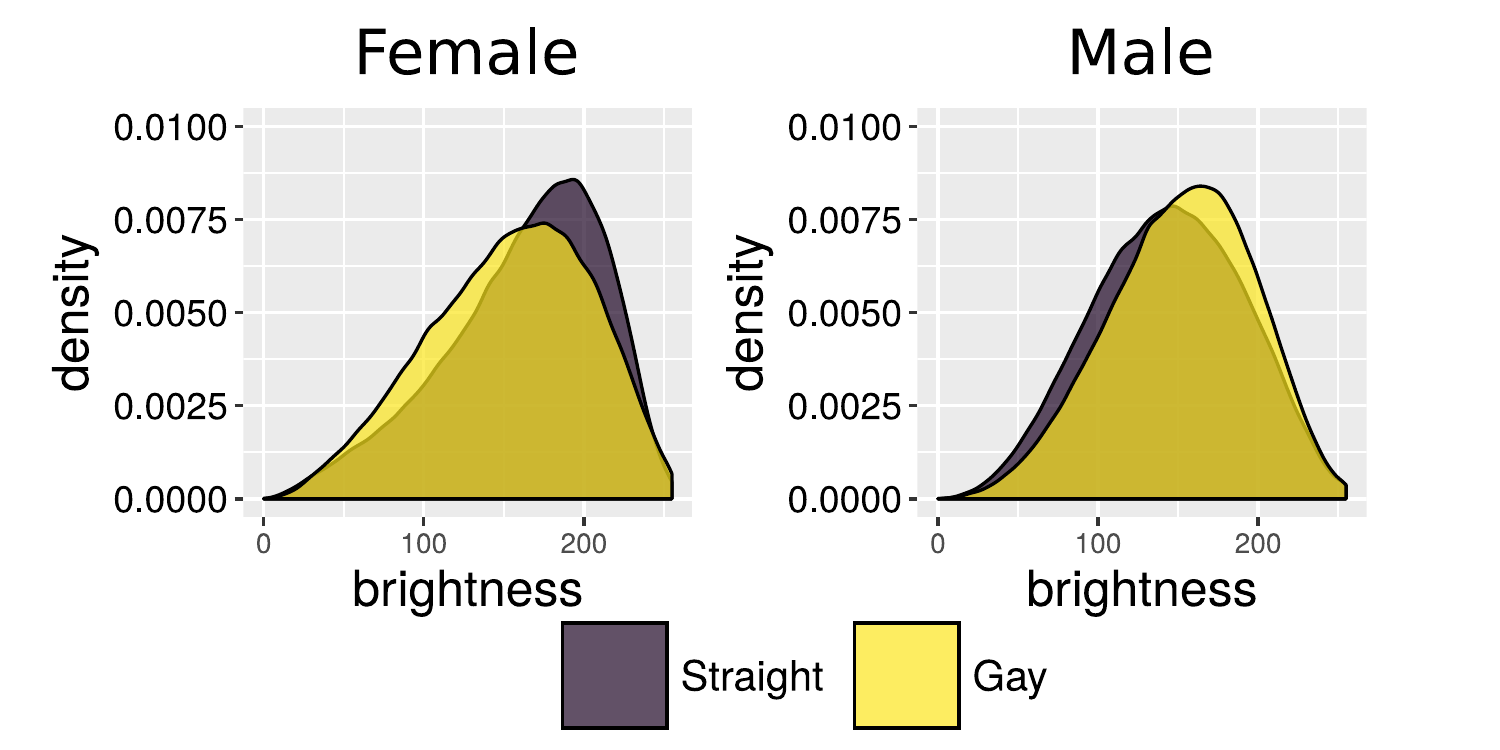}}%
	\caption{Brightness}%
	\label{fig:brightness_density}%
\end{figure*}

\label{app:appendix2}
\begin{figure*}[!htbp]
	\centering
	\centerline{\includegraphics{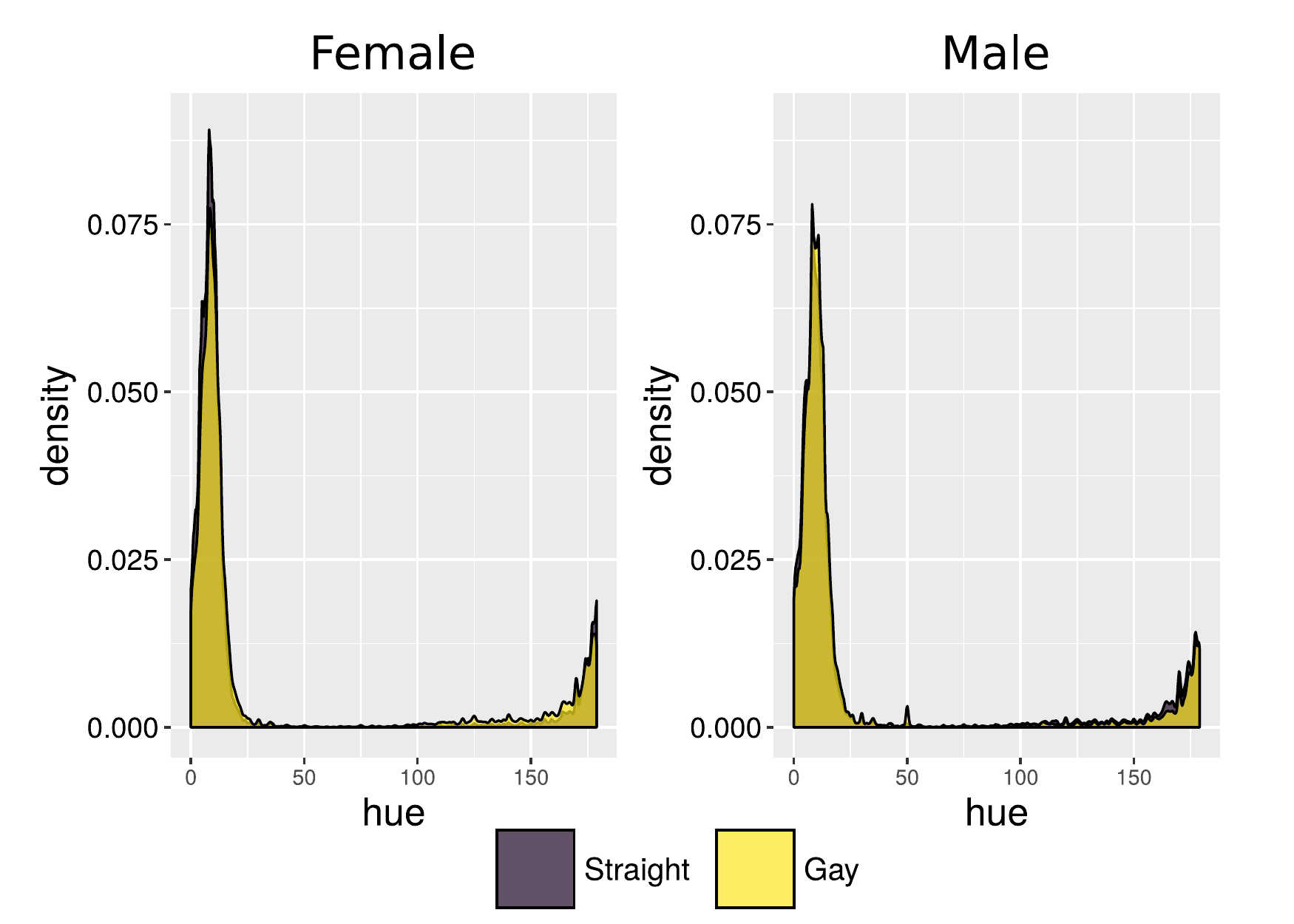}}
	\caption{Hue}%
	\label{fig:hue_density}%
\end{figure*}
Figures~\ref{fig:hue_density}, \ref{fig:saturation_density} and \ref{fig:brightness_density} plot the hue, saturation and brightness distributions for the highly blurred facial images used by Model~3 (Section \ref{model:blurred}).

To generate these distributions each 5$\plh$5 pixel blurred image was converted into the HSV (Hue, Saturation and Value) colour space \cite{foley}. Then the number of occurrences across all the images for a particular value were summed to create a density distribution. For saturation and brightness, values have an intensity between 0 and 255. The hues are mapped to a scale from 0 to 180 which wraps around at each end.

Each figure has two plots, one for females and one for males. Each plot shows the distribution of the colour metric in each category (Straight and Gay).

%%%%%%%%%%%%%%%%%%%%%%%%%%%%%%%%%%%%%%%%%%%%%%%%%
%%%%%%%%%%%%%%%%%%%%%%%%%%%%%%%%%%%%%%%%%%%%%%%%%

%\section{Summary}
%\label{sec:appendix1:summary}

%As always, provide a summary at the end.

%%%%%%%%%%%%%%%%%%%%%%%%%%%%%%%%%%%%%%%%%%%%%%%%%
%%%%%%%%%%%%%%%%%%%%%%%%%%%%%%%%%%%%%%%%%%%%%%%%%
% any further appendices you decide to write (create appropriate directory structure, as with chapters)
%\include{appendices/acronyms/main}
%\include{appendices/symbols/main}
%\include{appendices/derived_publications/main}

%%%%%%%%%%%%%%%%%%%%%%%%%%%%%%%%%%%%%%%%%%%%%%%%%
%%%%%%%%%%%%%%%%%%%%%%%%%%%%%%%%%%%%%%%%%%%%%%%%%

\cleardoublepage
\ifpdf
\phantomsection
\fi
%\label{index}
%\addcontentsline{toc}{chapter}{Index}
%\printindex

%%%%%%%%%%%%%%%%%%%%%%%%%%%%%%%%%%%%%%%%%%%%%%%%%
%%%%%%%%%%%%%%%%%%%%%%%%%%%%%%%%%%%%%%%%%%%%%%%%%

\end{document}